%% file: main.tex
\documentclass[conference]{preamble/IEEEtran}
\input{preamble/settings.tex}
\graphicspath{{./figures/images/}}
\begin{document}
\title{AK: Attentive Kernel for Information Gathering}
\input{preamble/authors.tex}
\maketitle
\begin{abstract}
  Robotic Information Gathering (RIG) relies on the uncertainty of a probabilistic model to identify critical areas for efficient data collection. Gaussian processes (GPs) with stationary kernels have been widely adopted for spatial modeling. However, real-world spatial data typically does not satisfy the assumption of \textit{stationarity}, where different locations are assumed to have the same degree of variability. As a result, the prediction uncertainty does not accurately capture prediction error, limiting the success of RIG algorithms. We propose a novel family of nonstationary kernels, named the Attentive Kernel (AK), which is simple, robust, and can extend any existing kernel to a nonstationary one. We evaluate the new kernel in elevation mapping tasks, where AK provides better accuracy and uncertainty quantification over the commonly used RBF kernel and other popular nonstationary kernels. The improved uncertainty quantification guides the downstream RIG planner to collect more valuable data around the high-error area, further increasing prediction accuracy. A field experiment demonstrates that the proposed method can guide an Autonomous Surface Vehicle (ASV) to prioritize data collection in locations with high spatial variations, enabling the model to characterize the salient environmental features.\footnote{Videos of the field/simulated experiments can be found at \url{https://weizhe-chen.github.io/attentive_kernels/}}
\end{abstract}
\IEEEpeerreviewmaketitle
\input{sections/1_introduction.tex}
\input{sections/2_literature.tex}
\input{sections/3_problem.tex}
\input{sections/4_methodology.tex}
\input{sections/5_experiments.tex}
\input{sections/6_conclusion.tex}
\balance
\bibliographystyle{unsrtnat}
\bibliography{references}
\onecolumn
\appendix
\input{sections/7_appendix.tex}
\end{document}

%% file: preamble/settings.tex
\usepackage{bm}
\usepackage{amssymb}
\usepackage{amsthm}
\usepackage{mathtools}
\usepackage{nicefrac}  

\usepackage[noend]{algpseudocode}
\usepackage{algorithm}
\usepackage{setspace}

\usepackage{booktabs}
\usepackage{multicol}
\usepackage{multirow}
\usepackage{tabularx}
\usepackage{etoolbox}
\newrobustcmd{\B}{\bfseries}  
\usepackage{siunitx}  
\usepackage{diagbox}
\usepackage{pifont}  
\usepackage{todo}  
\usepackage{color}
\usepackage{xcolor}
\definecolor{red}{RGB}{200,50,50}
\definecolor{green}{RGB}{50,200,50}
\definecolor{blue}{RGB}{50,50,200}
\definecolor{orange}{RGB}{243,147,82}
\definecolor{purple}{RGB}{150,100,250}
\definecolor{yellow}{RGB}{231,198,100}
\definecolor{gray}{RGB}{98,95,75}
\definecolor{white}{RGB}{255,255,255}
\usepackage{lipsum}
\usepackage{times}

\makeatletter
\newcommand*{\T}{{\mathpalette\@transpose{}} }
\newcommand*{\@transpose}[2]{\raisebox{\depth}{$\m@th#1\intercal$}}
\makeatother
\DeclarePairedDelimiterX{\norm}[1]{\lVert}{\rVert_{2}}{#1}

\DeclareMathOperator*{\argmin}{arg\,min}
\newcommand{\dataset}{{\mathbb{D}}}
\newcommand{\real}{\mathbb{R}}
\makeatletter
\newcommand\thefontsize{The current font size is: \f@size pt}
\makeatother

\makeatletter
\let\NAT@parse\undefined
\makeatother
\usepackage[numbers,sort&compress,square,comma]{natbib}  
\usepackage{hyperref}
\hypersetup{
  colorlinks=true,
  urlcolor=blue,
}
\usepackage{cleveref}  
\theoremstyle{definition}
\newtheorem{problem}{Problem}
\theoremstyle{definition}

\theoremstyle{definition}
\newtheorem{definition}{Definition}
\theoremstyle{definition}

\theoremstyle{definition}

\crefname{lemma}{Lemma}{Lemmas}
\crefname{proposition}{Proposition}{Propositions}
\crefname{definition}{Definition}{Definitions}
\crefname{theorem}{Theorem}{Theorems}
\crefname{conjecture}{Conjecture}{Conjectures}
\crefname{corollary}{Corollary}{Corollaries}
\crefname{section}{Section}{Sections}
\crefname{appendix}{Appendix}{Appendices}
\crefname{figure}{Fig.}{Figs.}
\crefname{equation}{Eq.}{Eqs.}
\crefname{table}{Table}{Tables}
\crefname{algocf}{Algorithm}{Algorithms}
\crefname{problem}{Problem}{Problems}
\usepackage{balance}  
\apptocmd{\sloppy}{\hbadness 10000\relax}{}{}

\usepackage{graphicx}
\usepackage[caption=false,font=footnotesize]{subfig}
\usepackage{wrapfig}  
\usepackage{tikz}
\usetikzlibrary{positioning,shapes,bayesnet,calc,backgrounds}
\newcommand*\circled[1]{\tikz[baseline=(char.base)]{\node[shape=circle,draw,inner sep=2pt] (char) {#1};}}
\makeatletter
\newif\if@showgrid@grid
\newif\if@showgrid@left
\newif\if@showgrid@right
\newif\if@showgrid@below
\newif\if@showgrid@above
\tikzset{%
  every show grid/.style={},
  show grid/.style={execute at end picture={\@showgrid{grid=true,#1}}},%
  show grid/.default={true},
  show grid/.cd,
  labels/.style={font={\sffamily\small},help lines},
  xlabels/.style={},
  ylabels/.style={},
  keep bb/.code={\useasboundingbox (current bounding box.south west) rectangle (current bounding box.north west);},
  true/.style={left,below},
  false/.style={left=false,right=false,above=false,below=false,grid=false},
  none/.style={left=false,right=false,above=false,below=false},
  all/.style={left=true,right=true,above=true,below=true},
  grid/.is if=@showgrid@grid,
  left/.is if=@showgrid@left,
  right/.is if=@showgrid@right,
  below/.is if=@showgrid@below,
  above/.is if=@showgrid@above,
  false,
}
\def\@showgrid#1{%
  \begin{scope}[every show grid,show grid/.cd,#1]
    \if@showgrid@grid
      \begin{pgfonlayer}{background}
        \draw [help lines]
        (current bounding box.south west) grid
        (current bounding box.north east);
        \pgfpointxy{1}{1}%
        \edef\xs{\the\pgf@x}%
        \edef\ys{\the\pgf@y}%
        \pgfpointanchor{current bounding box}{south west}
        \edef\xa{\the\pgf@x}%
        \edef\ya{\the\pgf@y}%
        \pgfpointanchor{current bounding box}{north east}
        \edef\xb{\the\pgf@x}%
        \edef\yb{\the\pgf@y}%
        \pgfmathtruncatemacro\xbeg{ceil(\xa/\xs)}
        \pgfmathtruncatemacro\xend{floor(\xb/\xs)}
        \if@showgrid@below
          \foreach \X in {\xbeg,...,\xend} {
            \node [below,show grid/labels,show grid/xlabels] at (\X,\ya) {\X};
          }
        \fi
        \if@showgrid@above
          \foreach \X in {\xbeg,...,\xend} {
            \node [above,show grid/labels,show grid/xlabels] at (\X,\yb) {\X};
          }
        \fi
        \pgfmathtruncatemacro\ybeg{ceil(\ya/\ys)}
        \pgfmathtruncatemacro\yend{floor(\yb/\ys)}
        \if@showgrid@left
          \foreach \Y in {\ybeg,...,\yend} {
            \node [left,show grid/labels,show grid/ylabels] at (\xa,\Y) {\Y};
          }
        \fi
        \if@showgrid@right
          \foreach \Y in {\ybeg,...,\yend} {
            \node [right,show grid/labels,show grid/ylabels] at (\xb,\Y) {\Y};
          }
        \fi
      \end{pgfonlayer}
      \fi
    \end{scope}
  }
  \makeatother
  \tikzset{every show grid/.style={show grid/keep bb}%
  }%

%% file: preamble/authors.tex
\author{
  \authorblockN{Weizhe Chen, Roni Khardon and Lantao Liu}
  \authorblockA{Indiana University, Bloomington, IN, USA, 47408\\
  Emails: \{chenweiz, rkhardon, lantao\}@iu.edu}
}

%% file: sections/1_introduction.tex
\section{Introduction}\label{sec:1}
Collecting informative data for effective modeling has been an active research topic in different domains, including active learning in machine learning~\cite{settles2009active}, optimal experimental design in statistics~\cite{atkinson1996usefulness}, and optimal placements in sensor networks~\cite{krause2008near}. {\em Robotic Information Gathering (RIG)} has recently received increasing attention due to its wide application, including environmental modeling and monitoring~\cite{flaspohler2019information,girdhar2014autonomous,hitz2017adaptive,ma2018data,manjanna2018heterogeneous,yu2014correlated,li2020exploration,popovic2017online,dunbabin2012robots,bai2021information}, 3D reconstruction and inspection~\cite{hollinger2013active,atanasov2014nonmyopic,kompis2021informed,zhu2021online}, search and rescue~\cite{meera2019obstacle,arora2017randomized}, autonomous exploration~\cite{arora2019multimodal,dang2018visual,thrun2004autonomous,shah2021rapid}, and system identification~\cite{capone2020localized,buisson2020actively,deisenroth2011pilco}. The defining element that distinguishes the aforementioned active information acquisition problems and RIG is the {robot embodiment}'s physical constraint -- we cannot ``teleport'' the robot to an arbitrary sampling location, and data must be collected sequentially along a trajectory. {\em Informative planning} seeks an action sequence or a policy that yields observations maximizing an {information-theoretic objective function} under the robot's motion and sensing budget constraints~\cite{hollinger2014sampling,best2019dec,chen2019pareto,popovic2020informative,schmid2020efficient,choudhury2018data,macdonald2019active,ghaffari2019sampling,zhang2020fsmi,schlotfeldt2021resilient,nishimura2021sacbp}. The objective is derived from the uncertainty of probabilistic models such as Gaussian processes (GPs)~\cite{ma2017informative,marchant2012bayesian,marchant2014bayesian,ghaffari2018gaussian,luo2018adaptive,ouyang2014multi,jang2020multi,krause2007nonmyopic,stachniss2009learning,popovic2020localization,lee2022trust}, Hilbert maps~\cite{ramos2016hilbert,senanayake2017bayesian,guizilini2019variational,senanayake2018automorphing}, occupancy grid maps~\cite{popovic2017online,popovic2020informative,saroya2021roadmap}, and Gaussian mixture models~\cite{tabib2019real,dhawale2020efficient,corah2019communication,meadra2018variable}.

\cref{fig:volcano_env} illustrates the workflow of a RIG system. The three major forces that drive RIG are (a) probabilistic models with well-calibrated uncertainty, (b) information-theoretic objective functions, and (c) informative planners. This work belongs to the first aspect: we aim to improve the uncertainty of GPs, which yields more informative objective functions for RIG. Such fundamental improvements can apply to any informative planner using any objective function.

\input{figures/tex/motivation}

Gaussian process regression (GPR) is one of the leading methods for mapping continuous spatiotemporal phenomena. Stationary kernels, \textit{e.g.}, the RBF kernel and the Mat\'ern family~\cite{rasmussen2005mit}, are commonly adopted in a GPR. However, real-world spatial data typically does not satisfy the assumption of {\em stationarity} -- i.e., that different locations have the same degree of variability. For instance, the environment in \cref{fig:volcano_env} has higher variability around the crater. As a result of such a mismatch, GPR with stationary kernels cannot portray the characteristic environmental features in detail. \cref{fig:volcano_rbf_prediction} shows the overs-moothed prediction. The prediction error and uncertainty are inconsistent (\textit{c.f.}, the circled region in \cref{fig:volcano_rbf_uncertainty} and \cref{fig:volcano_rbf_error}), leading to degraded performance if used with RIG.

There is extensive work on GPs with nonstationary data (\cref{sec:nonstat}). However, as shown in our experiments, prior work leaves room for improvement. The challenge is that nonstationary models are often too flexible to be trained. To address this, we propose a novel family of nonstationary kernels named the \emph{Attentive Kernel (AK)}. The main ideas behind the AK are limiting the nonstationary model to \emph{select} among a fixed set of correlation scales and masking out data across sharp transitions by {\em selecting} subsets of relevant data for each prediction. The ``soft'' selection process is learned from the data. \cref{fig:volcano_ak_prediction} shows GPR prediction with the AK on the same dataset. As highlighted by the arrows, the AK depicts the environment at a finer granularity. \cref{fig:volcano_ak_uncertainty} and \cref{fig:volcano_ak_error} show that the AK allocates higher uncertainty to the high-error area.

\textbf{Contributions.}
The main contribution of this paper is in designing the AK and evaluating its suitability for RIG. We present an extensive evaluation on elevation mapping tasks in several natural environments that exhibit a range of nonstationary features, comparing the AK to the stationary RBF kernel and the leading nonstationary kernels: the Gibbs kernel~\cite{gibbs1997bayesian,pacriorek2003nonstationary,remes2017nonstationary,heinonen2016nonstationary,lang2007adaptive,plagemann2008learning,plagemann2008nonstationary,remes2018neural} and Deep Kernel Learning (DKL)~\cite{wilson2016deep,calandra2016manifold}. The results show a significant advantage of the AK across passive learning, standard active learning, and RIG. We also present a field experiment to demonstrate the behavior of the proposed method in a real-world environment, where the prediction uncertainty of the AK guides an Autonomous Surface Vehicle (ASV) to identify essential sampling locations and collect valuable data rapidly. Last but not least, we release the code for reproducibility\footnote{\url{https://github.com/weizhe-chen/attentive_kernels}} and a software library for facilitating future research on RIG\footnote{\url{https://pypolo.readthedocs.io/}}.

%% file: figures/tex/motivation.tex
\begin{figure}[t]%
  \centering%
  \subfloat[Ground-Truth Environment\label{fig:volcano_env}]{%
  \includegraphics[width=0.6\linewidth]{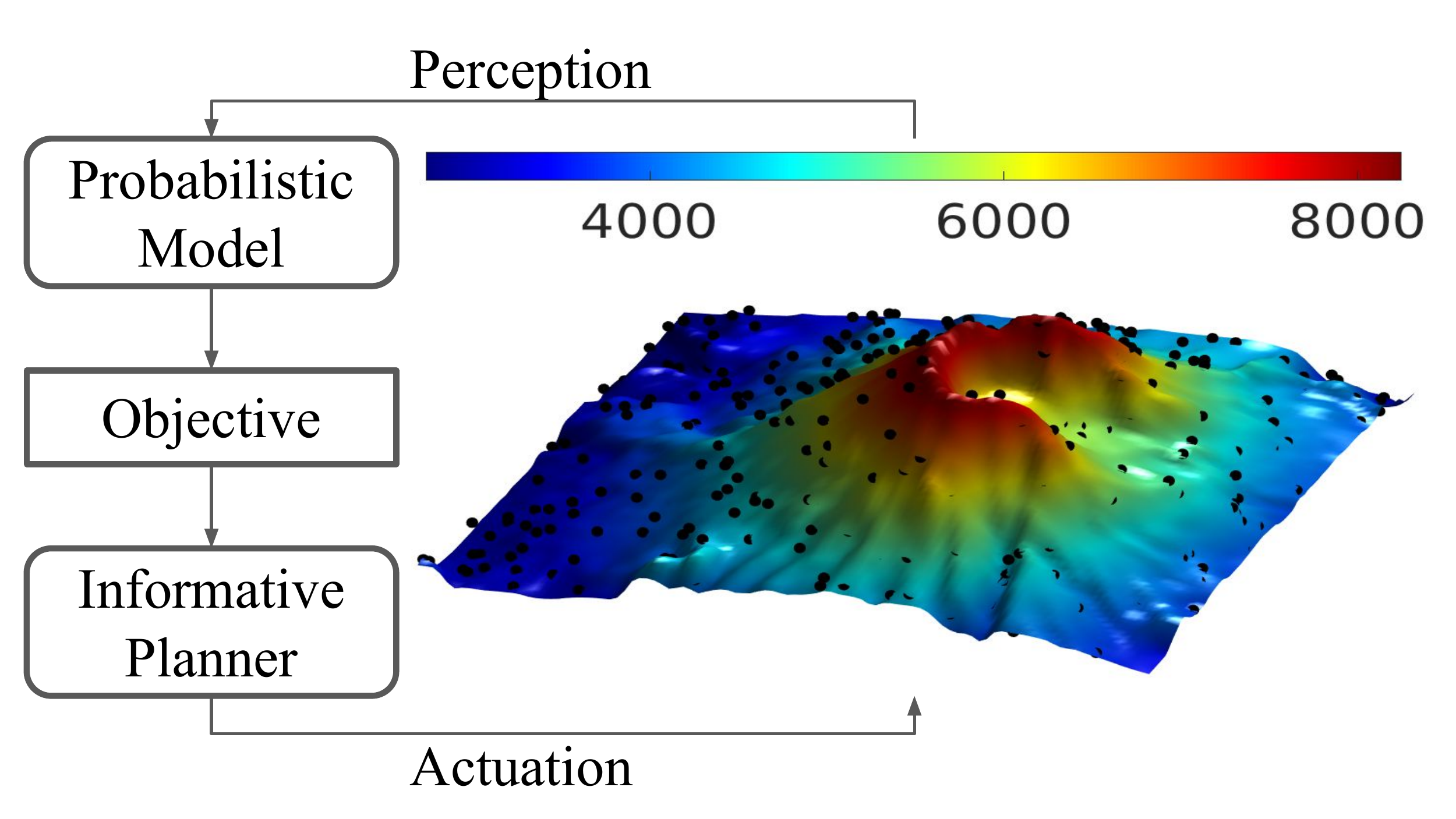}}\\
  \subfloat[Prediction of RBF\label{fig:volcano_rbf_prediction}]{%
    \resizebox{0.3\linewidth}{!}{%
      \includegraphics[width=0.5\linewidth,height=0.309\linewidth]{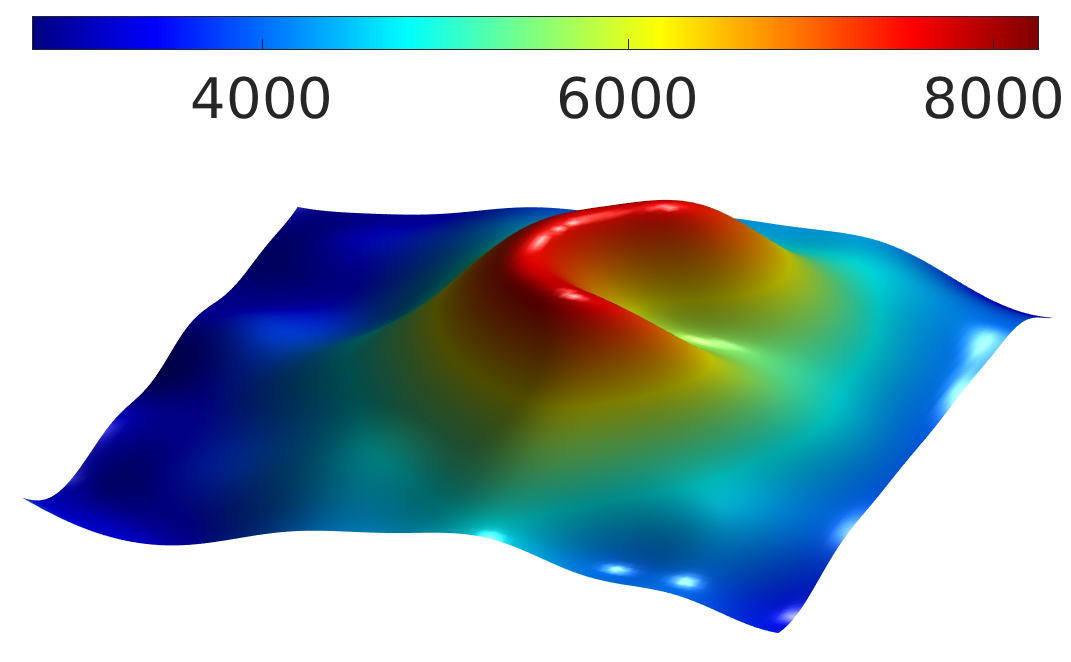}%
    }%
  }%
  \subfloat[Uncertainty of RBF\label{fig:volcano_rbf_uncertainty}]{%
    \resizebox{0.3\linewidth}{!}{%
      \begin{tikzpicture}%
        \node(a){\includegraphics[width=0.5\linewidth,height=0.309\linewidth]{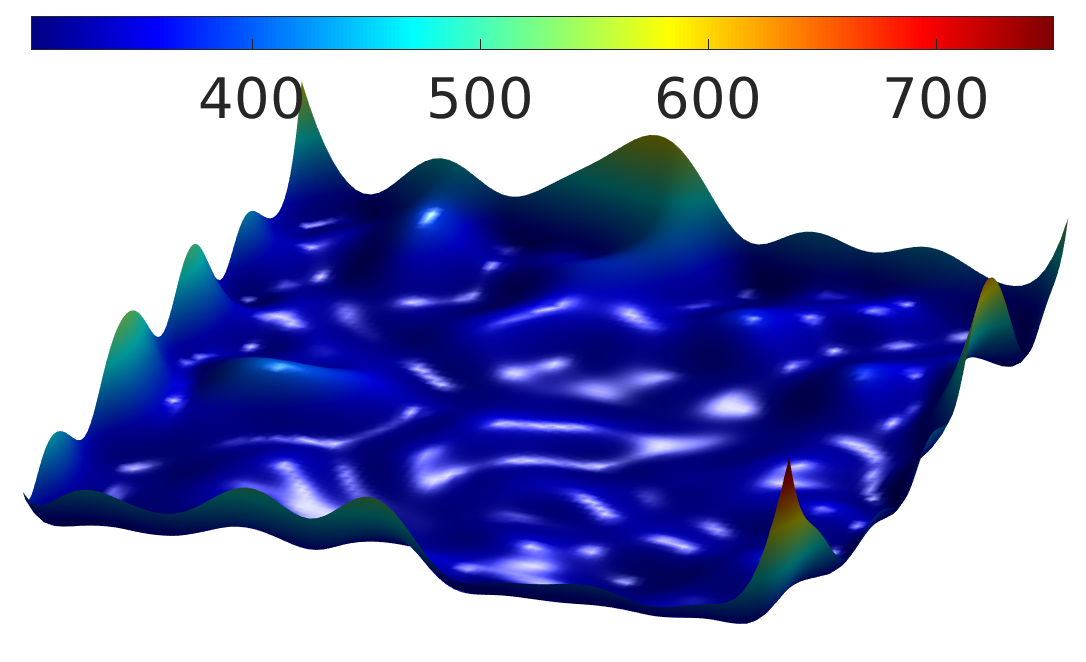}};%
        \node at(a.center)[draw,orange,line width=2,ellipse,minimum width=50,minimum height=50,rotate=0,yshift=-5,xshift=12]{};%
      \end{tikzpicture}%
    }%
  }%
  \subfloat[Error of RBF\label{fig:volcano_rbf_error}]{%
    \resizebox{0.3\linewidth}{!}{%
      \begin{tikzpicture}%
        \node(a){\includegraphics[width=0.5\linewidth,height=0.309\linewidth]{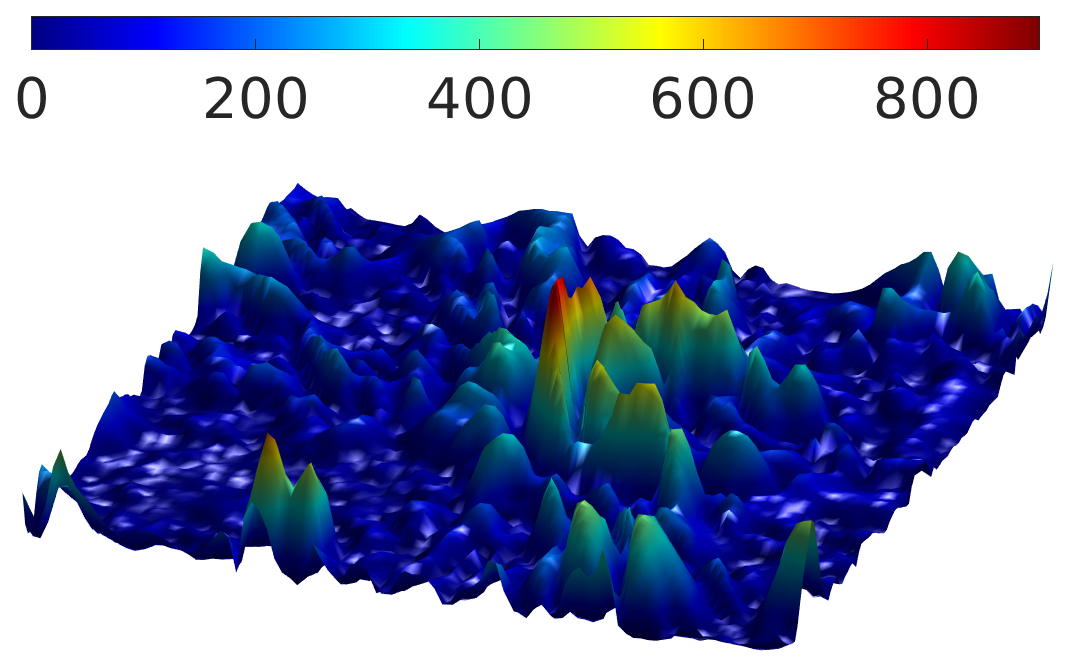}};%
        \node at(a.center)[draw,orange,line width=2,ellipse,minimum width=50,minimum height=50,rotate=0,yshift=-5,xshift=12]{};%
      \end{tikzpicture}%
    }%
  }\\%
  \subfloat[Prediction of AK\label{fig:volcano_ak_prediction}]{%
    \resizebox{0.3\linewidth}{!}{%
      \begin{tikzpicture}%
        \node(a){\includegraphics[width=0.5\linewidth,height=0.309\linewidth]{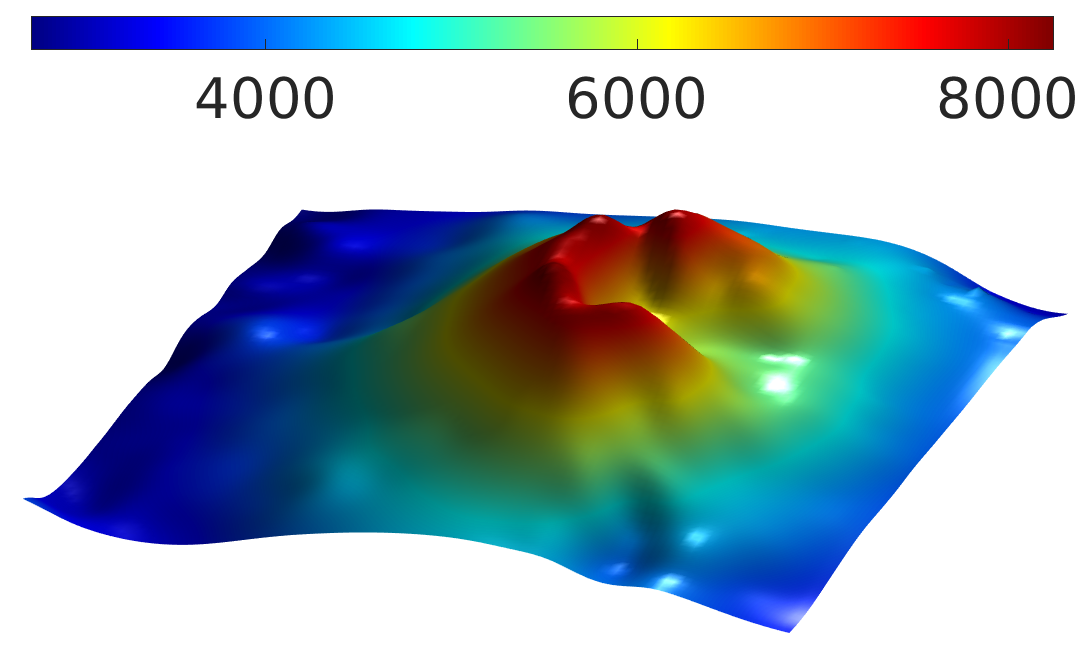}};%
        \draw [orange,-stealth,line width=1](0.5,0.8) -- (0.4,0.5);%
        \draw [orange,-stealth,line width=1](-0.1,-1.2) -- (0.2,-1.1);%
      \end{tikzpicture}%
    }%
  }%
  \subfloat[Uncertainty of AK\label{fig:volcano_ak_uncertainty}]{%
    \resizebox{0.3\linewidth}{!}{%
      \begin{tikzpicture}%
        \node(a){\includegraphics[width=0.5\linewidth,height=0.309\linewidth]{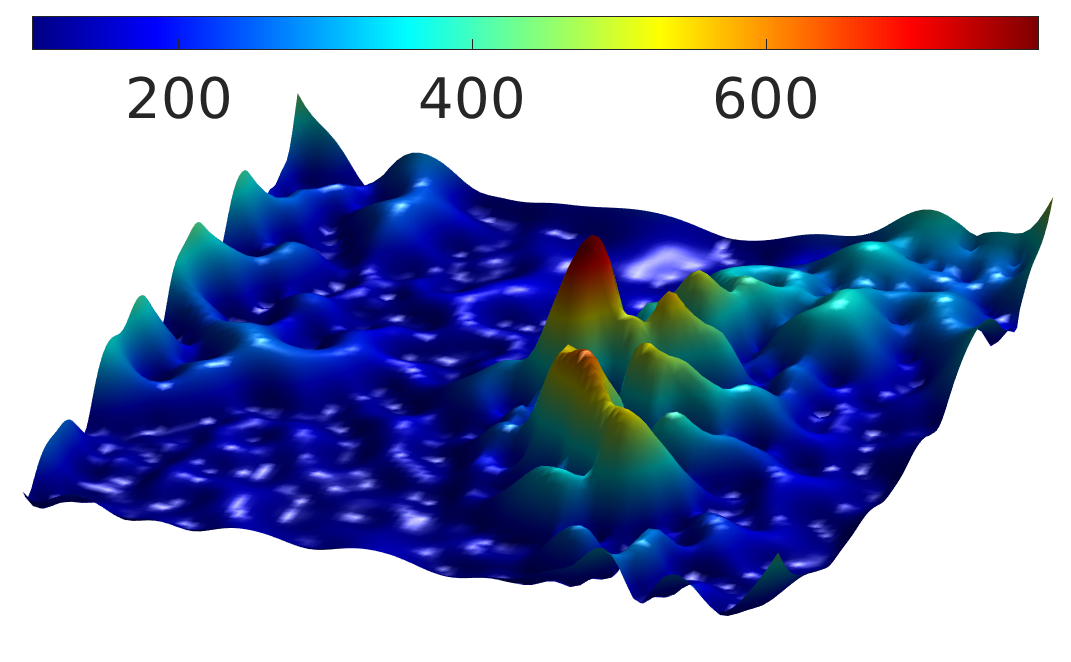}};%
        \node at(a.center)[draw,orange,line width=2,ellipse,minimum width=50,minimum height=50,rotate=0,yshift=-5,xshift=12]{};%
      \end{tikzpicture}%
    }%
  }%
  \subfloat[Error of AK\label{fig:volcano_ak_error}]{%
    \resizebox{0.3\linewidth}{!}{%
      \begin{tikzpicture}%
        \node(a){\includegraphics[width=0.5\linewidth,height=0.309\linewidth]{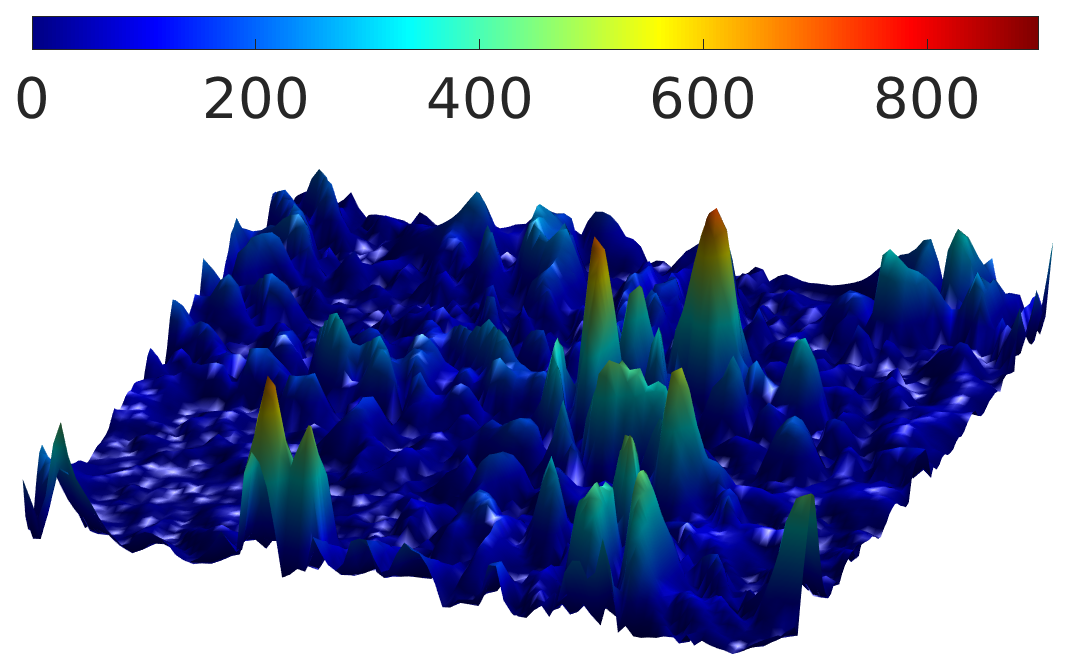}};%
        \node at(a.center)[draw,orange,line width=2,ellipse,minimum width=50,minimum height=50,rotate=0,yshift=-5,xshift=12]{};%
      \end{tikzpicture}%
    }%
  }%
  \caption{\textbf{Comparison of GPR models with RBF kernel and the AK in terrain mapping}. The color indicates elevation, and black dots are training samples. The AK portrays the salient environmental features in more detail and assigns higher uncertainty to the high-error area.}\label{fig:1}\vspace{-20pt}%
\end{figure}
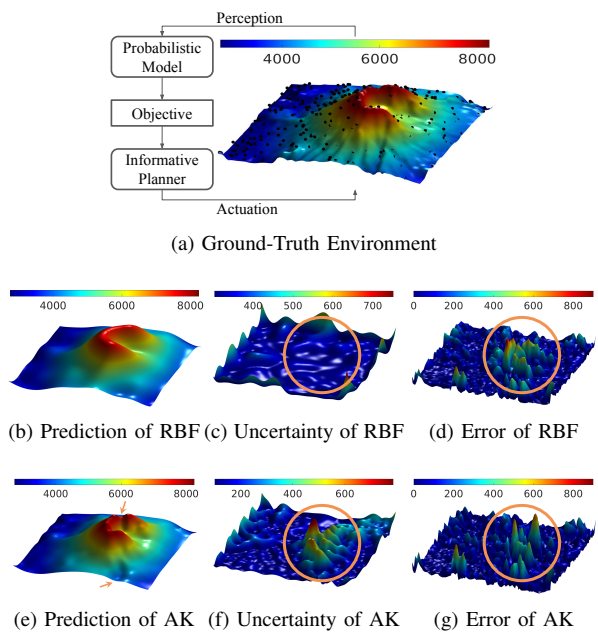

%% file: sections/2_literature.tex
\section{Related Work}\label{sec:2}
\subsection{Robotic Information Gathering}
Research on RIG mainly revolves around the following three aspects: probabilistic models, information-theoretic objective functions, and informative planning algorithms.

\textbf{Probabilistic Models and Objectives.}
Models are primarily discussed in coordinating multiple robots and improving computational efficiency. \citet{jang2020multi} apply the distributed GPs~\cite{deisenroth2015distributed} to decentralized multi-robot online active sensing. \citet{ma2017informative} and \citet{stachniss2009learning} use sparse GPs to alleviate the computational burden. Mixture models~\cite{rasmussen2001infinite} have been applied to divide the workspace into smaller parts for multiple robots to model an environment simultaneously~\cite{luo2018adaptive,ouyang2014multi}. The early work by \citet{krause2007nonmyopic} is highly related to our work. They use a spatially varying linear combination of localized stationary processes to model the nonstationary pH values. The weight of each local GP is the normalized predictive variance at the test location. This idea is similar to the lengthscale-selection idea in \cref{sec:lengthscale_selection}. The main difference is that they manually partition the workspace while our model learns a weighting function from data. To the best of our knowledge, this paper is the first to discuss the influence on RIG performance brought by the uncertainty quantification capability of probabilistic models. Research on information-theoretic objective functions is dedicated to addressing the computational bottleneck~\cite{charrow2015icra,charrow2015rss,zhang2020fsmi}.

\textbf{Informative Planning.}
Early works on informative planning propose various \textit{recursive greedy} algorithms that provide performance guarantee by exploiting the \textit{submodularity} property of the objective function~\cite{singh2007efficient,meliou2007nonmyopic,binney2013optimizing}. Planners based on dynamic programming lift the assumption of the objective function at the expense of higher computational complexity~\cite{low2009information,cao2013multi}. These methods solve combinatorial optimization problems in discrete domains, thus scaling poorly in the problem size. To develop efficient planners in \textit{continuous} space with motion constraints, \citet{hollinger2014sampling} introduce sampling-based informative motion planning, which is further developed to online variants~\cite{schmid2020efficient, ghaffari2019sampling}. Monte Carlo tree search planners are conceptually similar to sampling-based informative planners~\cite{kantaros2021sampling,schlotfeldt2018anytime} and have recently garnered great attention~\cite{arora2019multimodal,best2019dec,morere2017sequential,chen2019pareto,flaspohler2019information}. Trajectory optimization is a solid competitor to sampling-based planners. Bayesian optimization~\cite{marchant2012bayesian,bai2016information,di2021multi} and evolutionary strategy~\cite{popovic2017online,popovic2020informative,hitz2017adaptive} are the two dominating methods in this realm. New frameworks of RIG, \textit{e.g.}, imitation learning~\cite{choudhury2018data}, are constantly emerging.

\subsection{Nonstationary Gaussian Processes and Kernels}\label{sec:nonstat}
GPs suffer from two significant limitations~\cite{rasmussen2001infinite}. The first one is the notorious cubic computational complexity of a vanilla implementation. Recent years have witnessed remarkable progress in solving this problem based on sparse GPs~\cite{quinonero2005unifying,hoang2015unifying,bui2017unifying}. The second drawback is that the covariance function is commonly assumed to be stationary, limiting the modeling flexibility. Developing nonstationary GP models that are easy to train is still an active open research problem. Ideas of handling nonstationarity can be roughly grouped into three categories: input-dependent lengthscale~\citep{gibbs1997bayesian,pacriorek2003nonstationary,lang2007adaptive,plagemann2008learning,plagemann2008nonstationary,heinonen2016nonstationary,remes2017nonstationary}, input warping~\citep{sampson92nonparametric,snoek2014input,calandra2016manifold,wilson2016deep,tompkins2020sparse,salimbeni2017doubly}, and the mixture of experts~\citep{rasmussen2001infinite,trapp2020deep}.

Input-dependent lengthscale provides excellent flexibility to learn different correlation scales at different input locations, but optimizing the lengthscale function is difficult~\cite{wang2020nonseparable}. Input warping is more widely applicable because it endows any stationary kernel with the ability to model nonstationarity by mapping the input locations to a distorted space and assuming stationarity holds in the new space. This approach has a tricky requirement: the mapping must be \emph{injective} to avoid undesirable folding of the space~\citep{sampson92nonparametric,snoek2014input,salimbeni2017doubly}. The Mixture of GP experts (MoGPE) uses a \emph{gating network} to allocate each data to a local GP that learns its hyperparameters from the assigned data. It typically requires Gibbs sampling~\cite{rasmussen2001infinite}, which can be slow. Hence, one might need to develop a faster approximation~\cite{nguyen2008local}. We view MoGPE as an orthogonal direction to other nonstationary GPs/kernels because any GP models can be treated as the experts.

The AK lies at the intersection of these three categories. In \cref{sec:lengthscale_selection}, we implement input-dependent lengthscale by weighting base kernels with different prefixed lengthscales at each location. Composing base kernels reduces the difficulty of learning a lengthscale function from scratch and makes our method compatible with any base kernel. In \cref{sec:instance_selection}, we augment the input with extra dimensions. We can view the augmentation as warping the input space to a higher-dimensional space, ensuring \emph{injectivity} by design. Combining these two ideas gives a conceptually similar model to MoGPE~\cite{rasmussen2001infinite} in that they both divide the space into multiple regions and learn localized hyperparameters. The idea of augmenting the input dimensions has been discussed~\cite{pfingsten2006nonstationary}. However, they treat the augmented vector as a latent variable and resort to Markov chain Monte Carlo (MCMC) for inference. The AK treats the augmentation as a deterministic function of the input, resulting in a more straightforward inference procedure, and can be used in MoGPE to build more flexible models.

In robotic mapping, another line of notable work on probabilistic models is the family of Hilbert maps~\cite{ramos2016hilbert,senanayake2017bayesian,guizilini2019variational}, which aims to alleviate the computational bottleneck of the GP occupancy maps~\cite{callaghan2012Gaussian} by projecting the data to another feature space and applying a logistic regression classifier in the new space. Since Hilbert maps are typically used for occupancy mapping~\cite{doherty2016probabilistic} and reconstruction tasks~\cite{guizilini2017learning}, related work also considers nonstationarity for better prediction~\cite{senanayake2018automorphing,tompkins2020online}.

%% file: sections/3_problem.tex
\section{Problem Statement}\label{sec:3}
Consider deploying a robot to build a map of an \emph{initially unknown} environment \emph{efficiently} using the \emph{sparse} sensing measurements of onboard sensors. For instance, when reconstructing a pollution distribution map, the environmental sensors can only measure the pollutant concentration in a point-wise sampling manner, yielding sparse measurements along the trajectory. Another scenario is to build a sizeable bathymetric map of the seabed. In such a vast space, depth measurements can be viewed as \emph{point measurements} even though the sensor might be a multi-beam sonar. Exhaustively sampling the whole environment is prohibitive, if not impossible; thus, one must develop adaptive planning algorithms to collect the most informative data for building an accurate model. This problem is RIG, \textit{a.k.a.} informative (path/motion) planning, or active/adaptive sensing.

\begin{problem}\label{prob:1}%
  The target environment is an unknown function $\mathtt{f}_{\text{env}}(\mathbf{x}):\real^{D}\mapsto\real$ defined over spatial locations $\mathbf{x}\in\real^{D}$. Let $\mathbb{T}\triangleq\{t\}_{t=0}^{T}$ be the set of decision epochs. A robot at state $\mathbf{s}_{t-1}\in\mathcal{S}$ takes an action $a_{t-1}\in\mathcal{A}$, arrives at the next state $\mathbf{s}_{t}$ following a transition function $p(\mathbf{s}_{t}\mid\mathbf{s}_{t-1},a_{t-1})$, and collects $N_{t}\in\mathbb{N}$ noisy measurements $\mathbf{y}_{t}\in\real^{N_{t}}$ at sampling locations $\mathbf{X}_{t}=[\mathbf{x}_{1},\dots,\mathbf{x}_{N_{t}}]^{\T}\in\real^{N_{t}\times{D}}$. We assume that the transition function is known and deterministic, and the robot state is observable. The robot maintains a probabilistic model built from all the data collected so far $\mathbb{D}_{t}=\{(\mathbf{X}_{i},\mathbf{y}_{i})\}_{i=1}^{t}$. The model provides predictive mean $\mathtt{\mu}_{t}:\real^{D}\mapsto\real$ and predictive variance $\mathtt{\nu}_{t}:\real^{D}\mapsto\real_{\geq{0}}$ functions. Let $\mathbf{x}^{\star}$ be a test/query location and $\mathtt{error}(\cdot)$ be an error metric. At each decision epoch $t\in\mathbb{T}$, our goal is to find sampling locations that minimize the expected error after updating the model with the collected data%
\begin{equation}\label{eq:problem_error}%
  \argmin_{\mathbf{X}_{t}}\mathtt{E}_{\mathbf{x}^{\star}}\left[\mathtt{error}\left(\mathtt{f}_{\text{env}}(\mathbf{x}^{\star}),\mathtt{\mu}_{t}(\mathbf{x}^{\star}),\mathtt{\nu}_{t}(\mathbf{x}^{\star})\right)\right].%
\end{equation}%
\end{problem}%
\cref{eq:problem_error} cannot be used as the objective function for a planner because the ground-truth function $\mathtt{f}_{\text{env}}$ is unknown. RIG bypasses this problem by optimizing a surrogate objective.

\begin{problem}\label{prob:2}%
  Assuming the same conditions as \cref{prob:1}, find \emph{informative} sampling locations that minimize an information-theoretic objective function $\mathtt{info}(\cdot)$, \textit{e.g.,} entropy:
\begin{equation}\label{eq:problem_info}%
\argmin_{\mathbf{X}_{t}}\mathtt{E}_{\mathbf{x}^{\star}}\left[\mathtt{info}\left(\mathtt{\nu}_{t}(\mathbf{x}^{\star})\right)\right].%
\end{equation}
\end{problem}

RIG implicitly assumes that solving \cref{prob:2} can also address \cref{prob:1} well. This assumption is valid when the model uncertainty is \emph{well-calibrated}. A model with well-calibrated uncertainty gives high uncertainty when the prediction error is significant and low uncertainty otherwise. When using GPR with the commonly used stationary kernels to reconstruct a real-world environment, the uncertainty is not well-calibrated because the assumption of stationarity does not hold. Specifically, high uncertainty is assigned to the less sampled areas, regardless of the prediction error (see \cref{fig:volcano_rbf_uncertainty,fig:1d_rbf,fig:RBF_uncertainty}). Our goal is to develop a kernel to improve the uncertainty quantification and prediction accuracy of GPR.

%% file: sections/4_methodology.tex
\section{Methodology}\label{sec:4}
A Gaussian process is a collection of random variables, any finite number of which have a joint Gaussian distribution~\citep{rasmussen2005mit}. We place a Gaussian process prior over the function $\mathtt{f}(\mathbf{x})\sim\mathcal{GP}(\mathtt{m}(\mathbf{x}),\mathtt{k}(\mathbf{x},\mathbf{x}'))$, which is specified by a mean function $\mathtt{m}(\mathbf{x})$ and a covariance function $\mathtt{k}(\mathbf{x},\mathbf{x}')$ (a.k.a. kernel). GPR assumes that observations are corrupted by additive Gaussian white noise $p(y|\mathbf{x})=\mathcal{N}(y|\mathtt{f}(\mathbf{x}),\sigma^2)$, with noise scale $\sigma$. This paper focuses on the kernel construction, independent of the inference method in GPs. Therefore, we skip the discussion of inference methods and use the standard maximization of marginal likelihood to optimize the model in our experiments.

\subsection{Attentive Kernel}
We propose the following kernel to deal with nonstationarity. At first glance, this looks like a heuristic composite kernel. However, the following sections will explain how we design this kernel from the first principles. In short, the kernel is distilled from a generative model called AKGPR that models nonstationary processes.

\begin{definition}[Attentive Kernel]
 Given two inputs $\mathbf{x},\mathbf{x}'\in\real^{D}$, vector-valued functions $\mathtt{\mathbf{w}}_{\bm{\theta}}(\mathbf{x}):\real^{D}\mapsto[0,1]^{M}$ and $\mathtt{\mathbf{z}}_{\bm{\phi}}(\mathbf{x}):\real^{D}\mapsto[0,1]^{M}$ parameterized by $\bm{\theta,\phi}$, an amplitude $\alpha$, and a set of $M$ base kernels $\{\mathtt{k}_{m}(\mathbf{x},\mathbf{x}')\}_{m=1}^{M}$, let $\bar{\mathbf{w}}=\nicefrac{\mathbf{w}_{\bm{\theta}}(\mathbf{x})}{\norm{\mathtt{\mathbf{w}}_{\bm{\theta}}(\mathbf{x})}}$, and $\bar{\mathbf{z}}=\nicefrac{\mathbf{z}_{\bm{\phi}}(\mathbf{x})}{\norm{\mathtt{\mathbf{z}}_{\bm{\phi}}(\mathbf{x})}}$. The Attentive Kernel is defined as
  \begin{equation}
    \mathtt{ak}(\mathbf{x},\mathbf{x}')=\alpha\bar{\mathbf{z}}^{\T}\bar{\mathbf{z}}'\sum_{m=1}^{M}\bar{w}_{m}  \mathtt{k}_{m}(\mathbf{x},\mathbf{x}')\bar{w}_{m}'.
  \end{equation}\label{def:attentive_kernel}\vspace{-10pt}%
\end{definition}

We learn parametric functions that map each input $\mathbf{x}$ to $\mathbf{w}$ and $\mathbf{z}$. $\bar{w}_{m}\bar{w}_{m}'$ gives \emph{similarity attention scores} to weight the set of base kernels $\{\mathtt{k}_{m}(\mathbf{x},\mathbf{x}')\}_{m=1}^{M}$. The inner product $\bar{\mathbf{z}}^{\T}\bar{\mathbf{z}}'$ defines a \emph{visibility attention score} to mask the kernel value. \cref{def:attentive_kernel} is generic in that any existing kernel can be the base kernel. To address nonstationarity, we choose the base kernels to be a set of stationary kernels with the same functional form but different lengthscales. Specifically, we use RBF kernels with $M$ evenly spaced lengthscales in the interval $[\ell_{\text{min}},\ell_{\text{max}}]$:
\begin{equation}
  \mathtt{k}_{m}(\mathbf{x},\mathbf{x}')=\mathtt{\exp}\left(-\frac{\norm{\mathbf{x}-\mathbf{x}'}^{2}}{2\ell_{m}^{2}}\right),m=1,\dots,M.
\end{equation}
Note that the lengthscales $\{\ell_{m}\}_{m=1}^{M}$ are prefixed constants rather than trainable variables. When applying the attentive kernel to a GPR, we optimize all the hyperparameters $\{\alpha,\bm{\theta,\phi},\sigma\}$ by maximizing the marginal likelihood, and make prediction in the standard way. 

\subsection{Two Types of Nonstationarity}\label{sec:two_type_nonstat}
\input{figures/tex/1d_rbf.tex}
The example in \cref{fig:1d_rbf} motivates us to consider using different lengthscales at different input locations. Ideally, we need a smaller lengthscale for partition\#3 and larger lengthscales for the others. In addition, we need to break the correlations among data points in different partitions. An ideal nonstationary model should handle these two types of nonstationarity.

\citet{gibbs1997bayesian} and \citet{pacriorek2003nonstationary} have shown how one can construct a valid kernel with input-dependent lengthscales, namely, a lengthscale {\em function}. The standard approach uses another GP to model the lengthscale function, which is then used in the kernel of a GP, yielding a hierarchical Bayesian model. Several papers have developed inference techniques for such models and demonstrated their use in some applications~\citep{lang2007adaptive,plagemann2008learning,plagemann2008nonstationary,heinonen2016nonstationary,remes2017nonstationary}. Recently, \citet{remes2018neural} have shown that modeling the lengthscale function using a neural network improves performance.

However, the parameter optimization of such models is sensitive to data distribution and parameter initialization and leaves room for improvement. To address this, we propose a new approach that \emph{avoids learning a lengthscale function explicitly}. Instead, every input location can (a) \emph{select} among a set of GPs with different predefined primitive lengthscales and (b) \emph{select} which training samples are used when making a prediction. This idea -- selecting instead of inferring a localized lengthscale -- avoids difficulties in prior work. These ideas are developed in the following sections.

\subsection{Lengthscale Selection}\label{sec:lengthscale_selection}
\input{figures/tex/sin_weight.tex}
Consider a set of $M$ independent GPs with a set of predefined primitive lengthscales $\{\ell_{m}\}_{m=1}^{M}$. Intuitively, if every input location can select a GP with an appropriate lengthscale to make prediction, the nonstationarity can be handled well. We can achieve this by an \emph{input-dependent} weighted sum
\begin{align}
  \mathtt{f}(\mathbf{x})&=\sum_{m}^{M}\mathtt{w}_{m}(\mathbf{x})\mathtt{g}_{m}(\mathbf{x}),\text{ where}\label{eq:mix_ell}\\
  \mathtt{g}_{m}(\mathbf{x})&\sim\mathcal{GP}(\mathtt{0},\mathtt{k}(\mathbf{x},\mathbf{x}'\rvert{\ell_{m}})).\label{eq:g_gp}
\end{align}
$\mathtt{w}_{m}(\mathbf{x})$ is the $m$-th output of a vector-valued weighting function $\mathbf{w}_{\bm{\theta}}(\mathbf{x})$ parameterized by $\bm{\theta}$.

Consider the extreme case where $\mathbf{w}=[\mathtt{w}_{1}(\mathbf{x}),\dots,\mathtt{w}_{M}(\mathbf{x})]^{\T}$ is an ``one-hot'' vector -- a binary vector with only one element being 1. In this case, $\mathbf{w}$ selects one GP, and hence one lengthscale, depending on the input location. Inference techniques such as Gibbs sampling or Expectation Maximization are often required for learning such discrete ``assignment'' parameters. We lift this requirement by a continuous relaxation:
\begin{equation}
  \mathbf{w}_{\bm{\theta}}(\mathbf{x})=\mathtt{\mathbf{softmax}}(\tilde{\mathbf{w}}_{\bm{\theta}}(\mathbf{x})).\label{eq:w_softmax}
\end{equation}
Here, $\mathbf{w}_{\bm{\theta}}(\mathbf{x})$ is differentiable \textit{w.r.t.} $\bm{\theta}$, which can be optimized by the marginal likelihood maximization via gradient ascent. Moreover, using ``soft'' weights has an advantage in modeling gradually changing nonstationarity, as shown in \cref{fig:sin_weight}. Note that dividing the function into several discrete regimes using extreme weights is not reasonable for such a gradually changing function.

\input{figures/tex/1d_weight.tex}
\cref{fig:1d_weight} shows that lengthscale selection better predicts the same dataset as in \cref{fig:1d_rbf}. We can effectively model both the jittery pattern in partition\#3 and the gentle variations in the other partitions. However, when facing abrupt changes, as shown in the circled area, the model can only select a very small lengthscale to accommodate the loose correlations among data. If samples near the abrupt change are not dense enough, a small lengthscale might bring us a consequential prediction error. The following section will explain how to handle abrupt changes using instance selection.

\subsection{Instance Selection}\label{sec:instance_selection}
Intuitively, an input-dependent lengthscale specifies each data point's neighborhood radius. Changing the radius cannot handle abrupt changes well because data sampled before and after an abrupt change should break their correlations even when they are close. We need to control the \emph{visibility} among samples: each sample learns only from other samples in the same subgroup. To this end, we associate each input with a \emph{membership vector} $\mathbf{z}$ and use the dot product between two membership vectors to control visibility. Two inputs are visible to each other when they hold similar memberships. Otherwise, their correlation will be masked out:
\begin{equation}\label{eq:instance_selection}
  \mathtt{k}([\mathbf{x},\mathbf{z}],[\mathbf{x}',\mathbf{z}'])=\mathbf{z}^{\T}\mathbf{z}'\mathtt{k}_{\text{base}}(\mathbf{x},\mathbf{x}').
\end{equation}
We can view this as input {\em dimension augmentation} where we append $\mathbf{z}$ to $\mathbf{x}$ but use a structured kernel in the joint space of $[\mathbf{x},\mathbf{z}]$. It is also helpful to understand the case of extreme-valued hard partitions. In this case, the dot product is equal to $1$  if $\mathbf{z}$ and $\mathbf{z}'$ are the same one-hot vector and is $0$ otherwise. That is to say, when two points have different memberships, Eq.\,\eqref{eq:instance_selection} masks the correlation to zero. In this way, we only use the subset of data points in the same group. To make the model more flexible and simplify the parameter optimization, we use soft memberships:
\begin{equation}\label{eq:z_softmax}
  \mathbf{z}_{\bm{\phi}}(\mathbf{x})=\mathtt{\mathbf{softmax}}(\tilde{\mathbf{z}}_{\bm{\phi}}(\mathbf{x})).
\end{equation}

\subsection{The AKGPR Model}\label{sec:akgpr}
Combining the two ideas, we get the AKGPR model. While this is not immediately apparent, we show below that this model can be separated into a standard GPR and the AK (\cref{def:attentive_kernel}). The generative model is as follows.
\begin{itemize}
  \item Generate $\mathbf{w}$ and $\mathbf{z}$ using Eq.\,\eqref{eq:w_softmax} and Eq.\,\eqref{eq:z_softmax}.
  \item Compute the kernel values using Eq.\,\eqref{eq:instance_selection}.
  \item Generate $g_{m}\triangleq\mathtt{g}_{m}(\mathbf{x})$ from the corresponding GP~\eqref{eq:g_gp}.
  \item Compute $\mathtt{f}(\mathbf{x})$ via Eq.\,\eqref{eq:mix_ell}.
  \item Generate $y$ from the Gaussian likelihood.
\end{itemize}

\textbf{Parameterization and Optimization.}
To instantiate an AKGPR model, we must specify the weighting function $\mathbf{w}_{\bm{\theta}}(\mathbf{x})$ and the membership function $\mathbf{z}_{\bm{\phi}}(\mathbf{x})$. Our implementation parameterizes these functions using a simple neural network with two hidden layers\footnote{This is an arbitrary choice for the sake of simplicity and modeling flexibility. Any other parametric functions should also work.}. We optimize all the parameters by maximizing the log marginal likelihood $\ln{p(\mathbf{y}\rvert{\sigma,\alpha,\bm{\theta,\phi}})}$ with a lower learning rate for the neural network parameters.

\input{figures/tex/1d_ak.tex}
\cref{fig:1d_ak} shows the prediction of the AKGPR on the example from \cref{fig:1d_rbf}. Now we can model the jittery part, the smooth parts, and the abrupt changes accurately. Compared to \cref{fig:1d_rbf} where uncertainty only depends on the density of samples, the uncertainty of AKGPR can better reveal the prediction error. The AKGPR puts more weight on the GPs with small lengthscales in partition\#3 and those with large lengthscales in other partitions. Note that the AKGPR switches the membership vector $\mathbf{z}$ in the circled area to mask the inter-partition correlations, which cannot be realized by lengthscale selection in \cref{fig:1d_weight}. Due to this modeling advantage, \cref{fig:1d_ak} is qualitatively better than \cref{fig:1d_weight}.

\subsection{Analysis}\label{sec:analysis}
\textbf{AKGPR is GPR with the AK.}
By the definition of GPs, the training function values $\mathbf{g}_{m}\triangleq[\mathtt{g}_{m}(\mathbf{x}_{1}),\dots,\mathtt{g}_{m}(\mathbf{x}_{N})]^{\T}$ and the test function value $g_{m}^{\star}\triangleq{\mathtt{g}}_{m}(\mathbf{x}^{\star})$ at arbitrary test input $\mathbf{x}^{\star}$ jointly follow a multivariate Gaussian distribution. Let $\mathbf{f}\triangleq[\mathtt{f}(\mathbf{x}_{1}),\dots,\mathtt{f}(\mathbf{x}_{N})]^{\T}$ and $f^{\star}\triangleq{\mathtt{f}(\mathbf{x}^{\star})}$. Aggregate the weights of $N$ training inputs into $\mathbf{w}_{m}\triangleq[\mathtt{w}_{m}(\mathbf{x}_{1}),\dots,\mathtt{w}_{m}(\mathbf{x}_{N})]^{\T}$ and denote $w_{m}^{\star}\triangleq{\mathtt{w}_{m}(\mathbf{x}^{\star})},\mathbf{W}_{m}=\mathtt{diag}(\mathbf{w}_{m})$. \cref{eq:mix_ell} implies that their joint vector is the sum of $M$ linearly transformed multivariate Gaussian variables, which also follows a multivariate Gaussian distribution:
\begin{align}
  \begin{bmatrix}
    \mathbf{f}\\
    f^{\star}
    \end{bmatrix}&=\sum_{m}^{M}\begin{bmatrix}
    \mathbf{W}_{m} & \mathbf{0}\\
    \mathbf{0}^{\T} & w_{m}^{\star}
    \end{bmatrix}\begin{bmatrix}
    \mathbf{g}_{m}\\
    g_{m}^{\star}
  \end{bmatrix}\sim\mathcal{N}
  \left(
    \mathbf{0},
    \begin{bmatrix}
      \mathbf{C} & \mathbf{c} \\
      \mathbf{c}^{\T} & c
    \end{bmatrix}
  \right),
  \text{ where }\notag\\
    \mathbf{C}&=\textstyle\sum_{m=1}^{M}\mathbf{W}_{m}\mathbf{K}_{m}\mathbf{W}_{m},\label{eq:self_covariance}\\
    \mathbf{c}&=\textstyle\sum_{m=1}^{M}\mathbf{W}_{m}\mathbf{k}_{m}w_{m}^{\star},\label{eq:cross_covariance}\\
    c&=\textstyle\sum_{m}^{M}w_{m}^{\star}k_{m}w_{m}^{\star}\label{eq:test_covariance}.
\end{align}
The kernel values are given by \cref{eq:instance_selection}. Now from \cref{eq:self_covariance,eq:cross_covariance,eq:test_covariance} we observe that AKGPR is equivalent to a GPR with the AK given in \cref{def:attentive_kernel}.

We normalize $\mathbf{w}$ and $\mathbf{z}$ with $\ell^{2}$\,-\,norm to ensure that the maximum kernel value (when $\mathbf{x}=\mathbf{x}'$) is $1$, and $\alpha$ is the only parameter that controls the amplitude. Otherwise, the interplay between the amplitude hyperparameter $\alpha$ and the scaling effect of the kernel makes the optimization unstable. The analysis above holds for the normalized versions of $\mathbf{x},\mathbf{z}$ as well.

\input{figures/tex/ak_diagram.tex}
\textbf{Computational Complexity.}
Kernel matrix computations are typically done in a batch manner to take advantage of the parallelism in linear algebra libraries. Fig.\,\ref{fig:ak_diagram} shows the computational diagram of the self-covariance matrix of an input matrix $\mathbf{X}\in\real^{N\times{D}}$ for the case where $\mathbf{\mathtt{w}}_{\theta}(\mathbf{x})$ and $\mathbf{\mathtt{z}}_{\phi}(\mathbf{x})$ are parameterized by the same function. The computation of a cross-covariance matrix and the case where $\mathbf{\mathtt{w}}_{\theta}(\mathbf{x})$ and $\mathbf{\mathtt{z}}_{\phi}(\mathbf{x})$ are parameterized separately are handled similarly. We first pass $\mathbf{X}$ to a two-hidden-layer neural network to get $\mathbf{W}\in\real^{N\times{M}}\text{ and }\mathbf{Z}\in\real^{N\times{M}}$. The computational complexity of this step is $\mathcal{O}(NDH+NH^2+NHM)$. Then, we compute a visibility masking matrix $\mathbf{O}=\mathbf{Z}\mathbf{Z}^{\T}$, which takes $\mathcal{O}(N^{2}M)$. After getting the pairwise distance matrix $\left(\mathcal{O}(N^{2}D)\right)$, we can compute the base kernel matrices using different lengthscales $\left(\mathcal{O}(N^{2})\right)$. The $m$-th kernel matrix is scaled by the outer-product matrix of the $m$-th column of $\mathbf{W}$, which takes $\mathcal{O}(N^{2}M)$. Finally, we sum up the scaled kernel matrices and multiply the result with the visibility masking matrix to get the AK matrix $\left(\mathcal{O}(N^{2}M)\right)$. We defer the discussion of the choices of network size $H$ and number of base kernels $M$ to the sensitivity analysis section~\cref{sec:sensitivity_and_ablation}. In short, these will be relatively small numbers, so the overall computational complexity is still $\mathcal{O}(N^{2}D)$.

\subsection{RIG with the AK}
\input{algorithms/system.tex}
Algorithm\,\ref{alg:rig_ak} puts AK in the context of RIG. The system requires the following input arguments: the maximum number of training data $N_{\text{max}}$, the initial kernel amplitude $\alpha$, the initial noise scale $\sigma$, a set of $M$ base kernels $\{\mathtt{k}_{m}(\mathbf{x},\mathbf{x}')\}_{m=1}^{M}$, functions $\mathbf{\mathtt{w}}_{\bm{\theta}}(\mathbf{x})$, $\mathbf{\mathtt{z}}_{\bm{\phi}}(\mathbf{x})$, and a sampling strategy. First, we need to compute the statistics to normalize the inputs $\mathbf{X}$ to be roughly in the range $[-1,1]$ and standardize the targets $\mathbf{y}$ to nearly have zero mean and unit variance~(line\,1). We can get these statistics from prior knowledge of the environment. The workspace extent is typically known, allowing the normalization statistics to be readily calculated. The target-value statistics can be rough estimates or computed from a pilot environment survey~\cite{kemna2018pilot}. Then, we instantiate an AK and a GPR with the given parameters~(lines\,2-3). At each decision epoch $t$, the sampling strategy proposes an informative waypoint based on the predictive entropy of the probabilistic model~(line\,6). The robot tracks the informative waypoint and collects samples along the trajectory~(line\,7). The new samples are normalized and standardized and then appended to the model's training set~(lines\,8-9). Finally, we maximize the log marginal likelihood for $N_{t}$ iterations~(line\,10). The robot repeats predicting~(hidden in line\,6), planning, sampling, and optimizing until the sampling budget is exceeded~(line\,5).

To reduce the number of parameters and increase training stability, in the experiments, we unify the two functions $\mathbf{\mathtt{w}}_{\bm{\theta}}(\mathbf{x})\triangleq\mathbf{\mathtt{z}}_{\bm{\phi}}(\mathbf{x})$ and use the same set of parameters $\bm{\theta}=\bm{\phi}$, namely, training only one shared network. We discuss the two-network implementation in the ablation study~(\cref{sec:sensitivity_and_ablation}). We also notice that Occam's razor effect in the marginal likelihood is insufficient for preventing overfitting when training nonstationary kernels for many iterations. However, the AK is less prone to overfitting than the Gibbs kernel and DKL (see \cref{sec:appendix_overfitting}). \citet{tompkins2020sparse} also raised this point in their overfitting analysis. The proper way of training GP models with nonstationary kernels is still an open research problem and has recently received increasing attention~\cite{ober2021promises,van2021feature}. Held-out validation or cross-validation is not suitable for RIG, which does not have access to a large amount of data and has a real-time constraint. We use a rule-of-thumb early-stopping training scheme that works well empirically. Specifically, we train the model on all the collected data $\mathbb{D}_{t}$ for $N_{t}$ iterations after collecting $N_{t}$ samples at the $t$-th epoch.

%% file: figures/tex/1d_rbf.tex
\begin{figure}[t]%
  \centering
  \subfloat{%
    \resizebox{\linewidth}{!}{%
      \includegraphics[width=\linewidth,trim={10 10 10 10},clip]{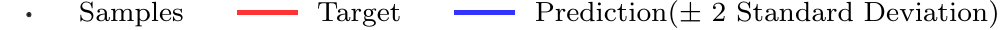}%
    }%
  }%
  \addtocounter{subfigure}{-1}\vspace{-8pt}\\%
  \subfloat[Wiggly prediction.\label{fig:1d_rbf_small_ell}]{%
    \resizebox{0.8\linewidth}{!}{%
      \begin{tikzpicture}%
        \node {\includegraphics[width=\linewidth,trim={7 7 7 7},clip]{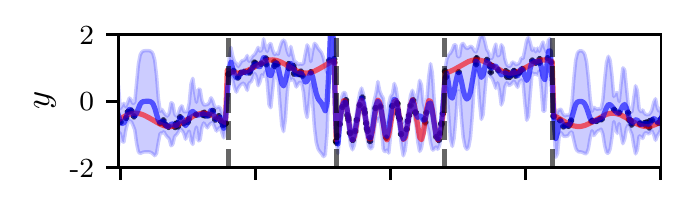}};%
        \node at (-1.95, 0.6) {\circled{1}};%
        \node at (-1.0, -0.6) {\circled{2}};%
        \node at (0.8, 0.6) {\circled{3}};%
        \node at (2.4, -0.5) {\circled{4}};%
        \node at (4.1, 0.65) {\circled{5}};%
      \end{tikzpicture}%
    }%
  }\vspace{-8pt}\\%
  \subfloat[Oversmoothed prediction.\label{fig:1d_rbf_large_ell}]{%
    \resizebox{0.8\linewidth}{!}{%
  \includegraphics[width=\linewidth,trim={7 7 7 7},clip]{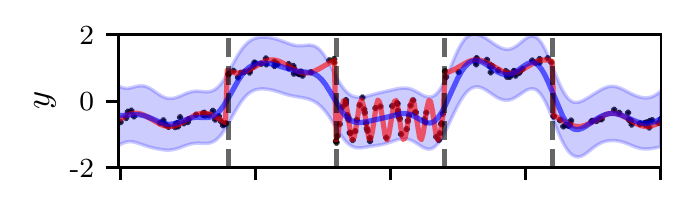}}}%
  \caption{\textbf{Learning a nonstationary function using GPR with RBF kernel}. The target function consists of five partitions separated by the dashed lines. The function changes drastically in partition\#3 and smoothly in the remaining partitions. The transitions between partitions are sharp. This simple function is challenging for a stationary kernel with a \textit{single} lengthscale hyperparameter. GPR with a stationary RBF kernel produces either the wiggly prediction shown in (a) or the over-smoothed prediction in (b). Note that, in (a), the prediction in the smooth regions is rugged, and the uncertainty is over-conservative when the training sample is sparse. The prediction in (b) only captures the general trend, and every input location seems equally uncertain.}\label{fig:1d_rbf}\vspace{-10pt}%
\end{figure}

%% file: figures/tex/sin_weight.tex
\begin{figure}[t]
  \centering%
  \subfloat{%
    \resizebox{0.8\linewidth}{!}{
      \begin{tikzpicture}%
        \node at(0.0, 0.0){\includegraphics[width=\linewidth,trim={7 7 7 7},clip]{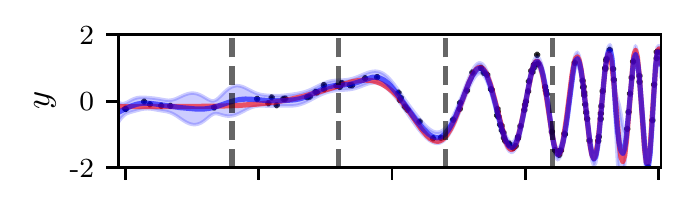}};%
        \node at(5.0, 0.0) {(a)};%
        \node at(0.1, -2.4){\includegraphics[width=0.98\linewidth,trim={7 7 7 7},clip]{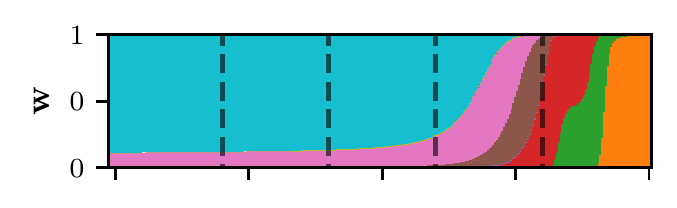}};%
        \node at(5.0, -2.4) {(b)};%
        \node at(0.7, -3.9){\includegraphics[width=0.85\linewidth,trim={10 10 10 10},clip]{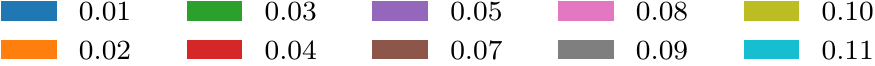}};%
      \end{tikzpicture}%
    }%
  }%
  \caption{\textbf{Learning $f(x)=x\sin(40x^{4})$ with soft lengthscale selection}. The $\mathbf{w}$-plot visualizes each input location's associated weighting vector $\mathbf{w}_{\bm{\theta}}(\mathbf{x})$. The more vertical length a color occupies, the higher weight we assign to the GP with the corresponding lengthscale. The learned weighting function gradually shift its weight from smooth GPs to bumpy ones.\vspace{-10pt}}\label{fig:sin_weight}%
\end{figure}

%% file: figures/tex/1d_weight.tex
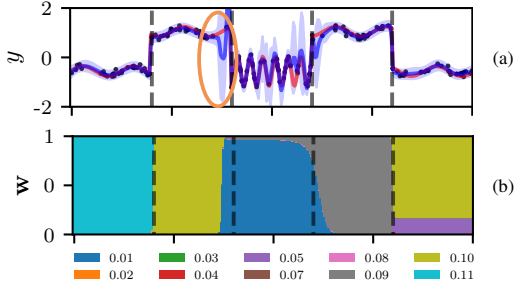
\begin{figure}[t]%
  \centering%
  \subfloat{%
    \resizebox{0.8\linewidth}{!}{
      \begin{tikzpicture}%
        \node at(0.0, 0.0){\includegraphics[width=\linewidth,trim={7 7 7 7},clip]{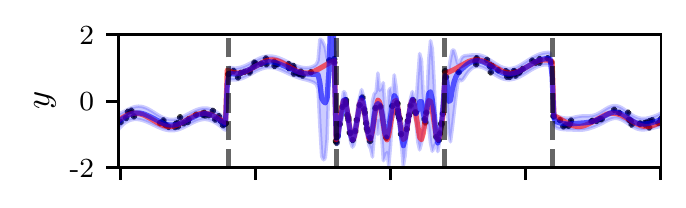}};%
        \node at(5.0, 0.0) {(a)};%
        \node at(a.center)[draw,orange,line width=2,ellipse,minimum width=20,minimum height=50,rotate=0,yshift=0pt,xshift=-10pt]{};%
        \node at(0.1, -2.4){\includegraphics[width=0.98\linewidth,trim={7 7 7 7},clip]{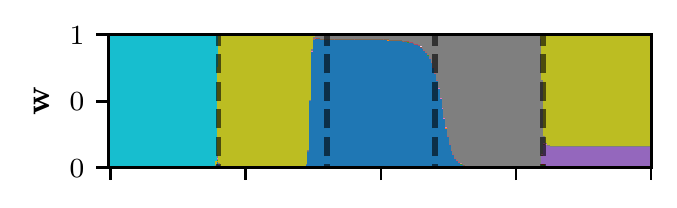}};%
        \node at(5.0, -2.4) {(b)};%
        \node at(0.7, -3.9){\includegraphics[width=0.85\linewidth,trim={10 10 10 10},clip]{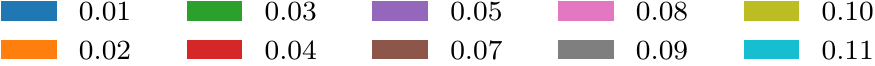}};%
      \end{tikzpicture}%
    }%
  }%
  \caption{\textbf{Learning the same function as in \cref{fig:1d_rbf} using lengthscale selection.}}\label{fig:1d_weight}\vspace{-10pt}%
\end{figure}

%% file: figures/tex/1d_ak.tex
\begin{figure}[t]%
  \centering%
  \subfloat{%
    \resizebox{0.8\linewidth}{!}{
      \begin{tikzpicture}%
        \node at(0.0, 0.0){\includegraphics[width=\linewidth,trim={7 7 7 7},clip]{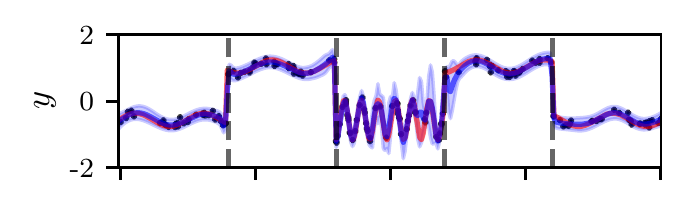}};%
        \node at(5.0, 0.0) {(a)};%
        \node at(a.center)[draw,orange,line width=2,ellipse,minimum width=20,minimum height=50,rotate=0,yshift=0pt,xshift=-10pt]{};%
        \node at(0.1, -2.4){\includegraphics[width=0.98\linewidth,trim={7 7 7 7},clip]{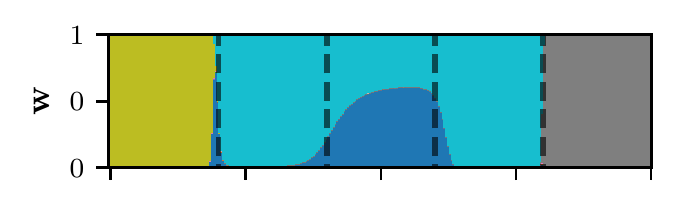}};%
        \node at(5.0, -2.4) {(b)};%
        \node at(0.7, -3.9){\includegraphics[width=0.85\linewidth,trim={10 10 10 10},clip]{demos/1d/legend_lengthscales.pdf}};%
        \node at(0.1, -5.4){\includegraphics[width=0.98\linewidth,trim={7 7 7 7},clip]{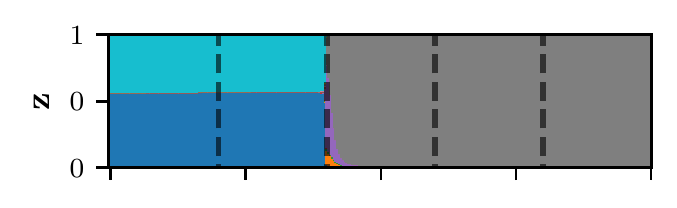}};%
        \node at(5.0, -5.4) {(c)};%
        \node at(0.7, -6.9){\includegraphics[width=0.65\linewidth,trim={10 10 10 10},clip]{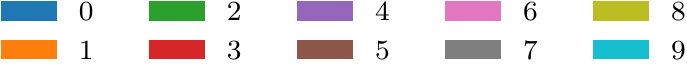}};%
      \end{tikzpicture}%
    }%
  }%
  \caption{\textbf{Learning the same function as in \cref{fig:1d_rbf} using AKGPR}. A weight/membership vector is visualized as a stack of bar plots produced by its elements. Different colors represent different lengthscales/dimensions of the weight/membership vector.}\label{fig:1d_ak}\vspace{-10pt}%
\end{figure}

%% file: figures/tex/ak_diagram.tex
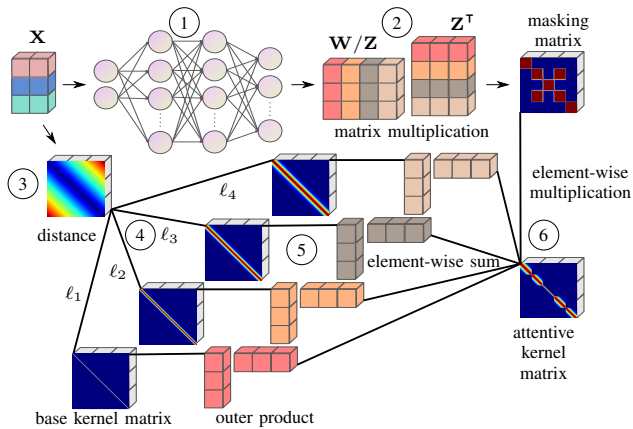
\begin{figure}[t]%
  \centering%
  \resizebox{\linewidth}{!}{%
    \begin{tikzpicture}\scriptsize%
      \node(a){\includegraphics[width=\linewidth]{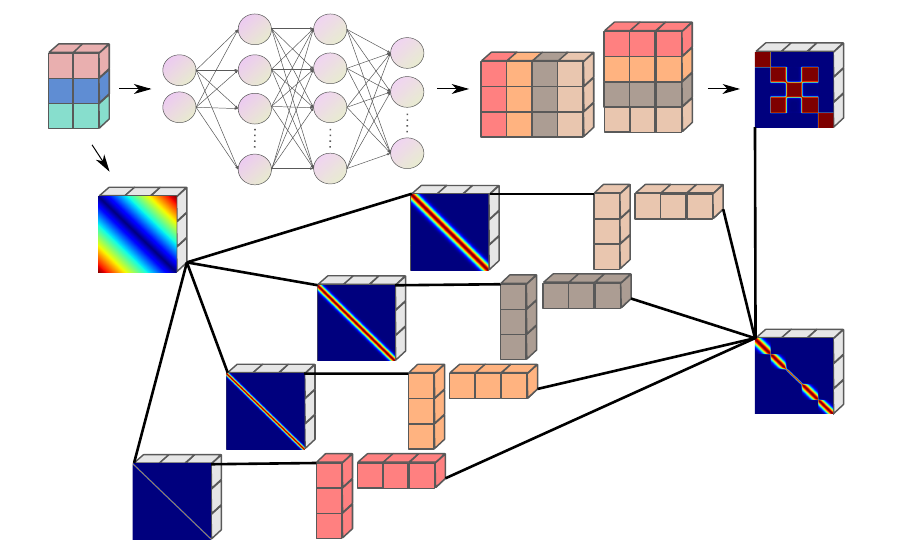}};%
      \node at (-3.6,  2.5) {$\mathbf{X}$};%
      \node at ( 0.7,  2.4) {$\mathbf{W}/\mathbf{Z}$};%
      \node at ( 2.2,  2.7) {$\mathbf{Z}^{\T}$};%
      \node[align=center] at ( 3.5, 2.6) {masking\\matrix};%
      \node at (-3.2, -0.2) {distance};%
      \node at (-3.1, -1.0) {$\ell_{1}$};%
      \node at (-2.5, -0.7) {$\ell_{2}$};%
      \node at (-1.8, -0.2) {$\ell_{3}$};%
      \node at (-1.0,  0.4) {$\ell_{4}$};%
      \node at (-2.7, -2.7) {base kernel matrix};%
      \node at (-0.5, -2.7) {outer product};%
      \node at (1.8, -0.55) {element-wise sum};%
      \node at ( 1.5,  1.2) {matrix multiplication};%
      \node[align=center] at ( 3.8,  0.5) {element-wise\\multiplication};%
      \node[align=center] at ( 3.3, -1.8) {attentive\\kernel\\matrix};%
      \node at (-1.6,  2.7) {\circled{1}};%
      \node at ( 1.3,  2.7) {\circled{2}};%
      \node at (-3.8,  0.5) {\circled{3}};%
      \node at (-2.2, -0.2) {\circled{4}};%
      \node at ( 0.0, -0.4) {\circled{5}};%
      \node at ( 3.3, -0.2) {\circled{6}};%
    \end{tikzpicture}%
  }%
  \caption{\textbf{Computational diagram of the AK}.}\label{fig:ak_diagram}\vspace{-10pt}%
\end{figure}

%% file: algorithms/system.tex
{\algrenewcommand\textproc{}
  \begin{algorithm}[t]\small%
    \setstretch{1.1}%
    \caption{RIG with The AK}\label{alg:rig_ak}%
    \hspace*{\algorithmicindent} \textbf{Arguments}: $N_{\text{max}},\alpha,\sigma,\{\mathtt{k}_{m}(\mathbf{x},\mathbf{x}')\}_{m=1}^{M}$\\
    \hspace*{\algorithmicindent}\hspace{48pt} $\mathbf{\mathtt{w}}_{\bm{\theta}}(\mathbf{x}),\mathbf{\mathtt{z}}_{\bm{\phi}}(\mathbf{x}),\mathtt{strategy}$
    \begin{algorithmic}[1] 
      \State compute normalization and standardization statistics
      \State $\mathtt{kernel\leftarrow\mathtt{AK}(\alpha,\{\mathtt{k}_{m}(\mathbf{x},\mathbf{x}')\}_{m=1}^{M},\mathbf{\mathtt{w}}_{\bm{\theta}}(\mathbf{x}),\mathbf{\mathtt{z}}_{\bm{\phi}}(\mathbf{x}))}$
      \State $\mathtt{model\leftarrow\mathtt{GPR}(\mathtt{kernel},\sigma)}$
      \State $t\leftarrow{0}$
      \While{$\mathtt{model}.N_{\text{train}}<N_{\text{max}}$}\Comment{sampling budget}
      \State $\mathtt{x_{\text{info}}\leftarrow\mathtt{strategy}(model)}$\Comment{informative waypoint}
      \State $\mathtt{\mathbf{X}_{t},\mathbf{y}_{t}\leftarrow{tracking\_and\_sampling}(x_{\text{info}})}$\Comment{$N_{t}$ samples}
      \State $\mathtt{\bar{\mathbf{X}}_{t},\bar{\mathbf{y}}_{t}\leftarrow{normalize\_and\_standardize}(\mathbf{X}_{t},\mathbf{y}_{t}})$
      \State $\mathtt{model.add\_data(\bar{\mathbf{X}}_{t},\bar{\mathbf{y}}_{t}})$
      \State $\mathtt{model.optimize}(N_{t})$\Comment{maximize marginal likelihood}
      \State $t\leftarrow{t+1}$
      \EndWhile
      \State \textbf{return} $\mathtt{model}$
    \end{algorithmic}
  \end{algorithm}
}

%% file: sections/5_experiments.tex
\section{Experiments}\label{sec:5}
We design our experiments to address the following questions. (Q1) Is the uncertainty quantification of AK better than its stationary counterpart and the nonstationary baselines? (Q2) Can we achieve better performance in active learning and RIG with the improved uncertainty quantification? (Q3) Is the performance of AK sensitive to the parameter settings? To answer Q1, we use random sampling experiments in Section~\ref{sec:random_sampling} to ensure that the sampling strategy does not bias the results. For Q2, we conduct both active learning~(\cref{sec:active_sampling}) and RIG experiments~(\cref{sec:informative_planning}) to disentangle the influence of the model and the planner. RIG relies on a planner that considers the physical constraints of the robot embodiment, while active learning is planner-agnostic. Finally, we address Q3 by evaluating the system under different configurations. 

\subsection{Experimental Setups}
\input{figures/tex/environments.tex}
\textbf{Environments.}
We have conducted extensive experiments in 4 environments that exhibit various nonstationary features. To shorten the discussion we present two environments here.  Complete results are given in \cref{sec:appendix_tables}. Fig.\,\ref{fig:envs} shows the two environments. In Fig.\,\ref{fig:N17E073}, the environment consists of a flat part, a mountainous area, and a rocky region with many ridges. The right part of Fig.\,\ref{fig:N47W124} varies drastically, and its left part is relatively flat.

\textbf{Probabilistic Model.}
We build a GPR with noisy training samples collected from the ground-truth digital elevation maps in all experiments. The GPR takes two-dimensional sampling locations as inputs and predicts the elevation. We first collect $50$ samples to have an initial optimization of the hyperparameters and compute the statistics to normalize the inputs and standardize the targets. 

\textbf{Sampling Strategies.}
We use different sampling strategies in the three sets of experiments. In random sampling experiments, we draw a sample uniformly at random at each decision epoch. In active sampling experiments, we evaluate the predictive uncertainty on a set of $1000$ randomly generated candidate locations and then sample from the location with the highest predictive entropy. While the AK can be plugged into any advanced informative planner for RIG, we use a simple planner to evaluate its performance. Specifically, in addition to the predictive entropy, this planner also computes the distances from these locations to the robot's position and normalizes the predictive entropy and distance to $[0, 1]$, respectively. Each candidate location's informativeness score is defined as the normalized entropy minus the normalized distance. This planner outputs the informative waypoint with the highest score. The robot moves to the waypoint via a tracking controller and samples along the path. Note that the number of collected samples $N_{t}$ varies at different decision epochs depending on the distance from the robot to the informative waypoint.

\textbf{Baselines.}
We compare AK with three popular kernels that have been recently shown to provide good performance. Among the three kernels, two are nonstationary, including the Gibbs kernel and DKL, and the third is the stationary RBF kernel widely used in RIG. Specifically, the Gibbs kernel extends the lengthscale to be any positive function of the input and degenerates to an RBF kernel when using a constant lengthscale function. Following \cite{remes2018neural} that showed improved results, the lengthscale function is modeled using a neural network instead of using a hierarchical process. DKL addresses nonstationarity through input warping. A neural network transforms the inputs to a feature space where stationary kernels are assumed to be sufficient. We use the same neural network for AK and DKL and change the output dimension to be one for the Gibbs kernel because it requires a scaler-valued lengthscale function.

\textbf{Metrics.}
We care about the prediction performance and whether the predictive uncertainty can effectively reflect the prediction error. Following standard practice in the GP literature, we use {\em standardized mean squared error (SMSE)} and {\em mean standardized log loss (MSLL)} to measure these quantities. SMSE is the mean squared error divided by the variance of test targets. With this standardization the trivial method of guessing the mean of the training targets has an SMSE of approximately $1$. To take the predictive uncertainty into account, one can evaluate the negative log probability, i.e., log loss, of a test target, $$-\ln{p(y^{\star}\rvert\dataset,\mathbf{x}^{\star})}=\frac{\ln(2\pi\nu)}{2}+\frac{(y^{\star}-\mu)^{2}}{(2\nu)},$$ where $\mu$ and $\nu$ are the mean and variance in the predictive distribution. MSLL standardizes the log loss by subtracting the loss obtained under the trivial model, which predicts using a Gaussian with the mean and variance of the training targets. The MSLL will be approximately zero for simple methods and negative for better methods. In the experiments, we also measured the root-mean-square error (RMSE), mean negative log-likelihood (MNLL), and mean absolute error (MAE). The results are consistent across different metrics, and the complete results with all metrics are given in \cref{sec:appendix_tables}. We report the mean and standard deviation of the metrics over $10$ runs of the experiments with different random seeds.

\input{figures/tex/metrics_random.tex}
\input{figures/tex/metrics_active.tex}
\input{figures/tex/metrics_myopic.tex}
\input{figures/tex/predictions.tex}
\input{figures/tex/field.tex}

\subsection{Random Sampling}\label{sec:random_sampling}
\cref{fig:N17E073_random_smse,fig:N47W124_random_smse} show that AK has better a prediction error than the other kernels with randomly sampled data. In these experiments, we are especially interested in the MSLL because a lower MSLL means that the model gives high uncertainty when its prediction is far from the test target, which can help active sampling and informative planning reduce the error faster. \cref{fig:N17E073_random_msll,fig:N47W124_random_msll} show that AK has a significant advantage in MSLL over the other methods. The Gibbs kernel also has some advantage over the other two methods, but the MSLL of DKL is almost the same as that of RBF.

\subsection{Active Sampling}\label{sec:active_sampling}
\cref{fig:N17E073_active_smse,fig:N47W124_active_smse} show that AK also has faster error reduction when the samples are actively collected. The AK can quickly identify the crucial areas that account for most of the error and sample more valuable data in those spots, leading to a significant gap in the final metrics. In \cref{fig:N17E073_active_msll,fig:N47W124_active_msll}, Gibbs and DKL improve over RBF in uncertainty quantification in the active sampling.

\subsection{Informative Planning}\label{sec:informative_planning}
Informative planning is a more challenging task than active learning because once the robot decides to visit an informative waypoint, it has to collect the samples along its trajectory. Given a fixed maximum number of samples, the number of decision epochs of RIG is much smaller than that of active sampling, which makes informed decisions more essential. \cref{fig:metrics_myopic} shows that AK is consistently leading in all the metrics with the informative planning strategy. The Gibbs kernel still has clear improvement over the RBF kernel in MSLL, but DKL falls short in these experiments.

\cref{fig:prediction} shows a snapshot of the prediction results of different methods after 400 samples, along with the ground-truth environment in the first row. The RBF kernel misses many environmental features that nonstationary kernels can capture. We observe the following behaviors by comparing the patterns in the uncertainty maps and error maps. Note that the error maps use the same color scale for ease of comparison across different methods. Each uncertainty map has its color scale -- red color only indicates relatively high uncertainty within the map. 
\begin{itemize}
  \item Regardless of the prediction errors, the RBF kernel gives the less sampled area higher uncertainty. The sampling path uniformly covers the space.
  \item The AK assigns higher uncertainty in the regions with more significant error. The sampling path focuses more on the complex region.
  \item The Gibbs kernel also has higher uncertainty in the rocky region but does not assign high uncertainty to the lower-right. Therefore, the sampling path concentrates on the upper-right corner and misses some high-error spots.
  \item When using DKL, the robot also samples the upper-right corner densely. The prediction error at the bottom of the map is the largest across different methods. However, DKL also places high uncertainty there.
\end{itemize}

\subsection{Sensitivity Analysis and Ablation Study}%
\label{sec:sensitivity_and_ablation}
We stress-test the AK under different parameter settings for sensitivity analysis and compare four variants of the AK for ablation study. We present the conclusions of the analysis here and provide full details of these tests in \cref{sec:appendix_sensitivity,sec:appendix_ablation}.

\textbf{Sensitivity Analysis.}
Increasing the number of base kernels or primitive lengthscales $M$ improves performance, albeit with a diminishing return and higher computational cost. Choosing $M$ in the range of $[5,10]$ is a good tradeoff between performance and computational efficiency. AK is not sensitive to the number of hidden units $H$ in the neural network as long as $H$ is not too small, e.g., only two units. Using smaller minimum lengthscales $\ell_{\text{min}}$ yields better performance, but the advantage of choosing a $\ell_{\text{min}}$ smaller than $0.01$ is negligible, so $0.01$ is an appropriate choice. AK is also robust to the setting of maximum lengthscale $\ell_{\text{max}}$ as long as it is not too small. After normalizing the input to be nearly in the range $[-1, 1]$, choosing $\ell_{\text{max}}$ between $[0.5, 1.0]$ is suitable. Overall, the AK is robust to various parameter settings.

\textbf{Ablation Study.}
The ablation study shows that lengthscale selection is necessary. Dropping it decreases the performance significantly. On the other hand, we do not observe a significant performance advantage from instance selection using the current training scheme. Nonetheless, as illustrated in Fig.\,\ref{fig:1d_ak}, we expect instance selection to provide better modeling of sharp transitions. Since instance selection improves the prediction only in a small region, the improvement might not be evident in the aggregated evaluation metrics. Using two separate neural networks does not provide an improvement. However, it deteriorates the uncertainty quantification in one environment. We conjecture that the two-network implementation might show its strength with a more refined approach to parameter training.

\subsection{Field Experiment}
We demonstrate the proposed AK in a RIG task -- active elevation/bathymetric mapping. We deploy an ASV with a single-beam sonar pointing downward to collect depth measurements~(\cref{fig:field_system}). The robot can localize itself by fusing the GPS and IMU data and actuate through the two thrusters. Our goal is to build an elevation map within the workspace shown in \cref{fig:field_workspace} with a small number of samples. From the satellite imagery, we can vaguely see interesting environmental features in the lower-left and upper-right corners of the workspace. \cref{fig:field_snapshot} shows a snapshot of the final model prediction, uncertainty, and sampling path. The prediction uncertainty is effectively reduced after sampling, and most of the samples are collected in critical regions with drastic elevation variations. Such a biased sampling pattern allows us to model the general trend of smooth regions with a small number of samples while capturing the characteristic environmental features at a fine granularity.

%% file: figures/tex/environments.tex
\begin{figure}[t]
  \centering
  \foreach \env in {N17E073,N47W124}{%
    \subfloat[\env\label{fig:\env}]{%
      \resizebox{0.5\linewidth}{!}{%
  \includegraphics[width=0.5\linewidth,height=0.209\linewidth,trim=8 8 8 8]{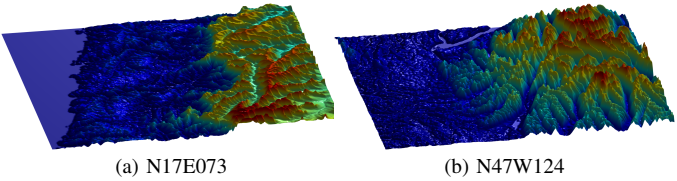}}}}%
  \caption{\textbf{Two of the environments used in the elevation mapping tasks}.}\label{fig:envs}%
\end{figure}

%% file: figures/tex/metrics_random.tex
\begin{figure}[t]
  \centering
  \hspace{10pt}\subfloat{%
    \resizebox{0.7\linewidth}{!}{%
      \includegraphics[width=\linewidth,trim=8 8 8 8]{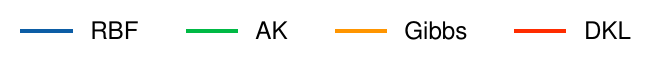}%
    }%
  }%
  \addtocounter{subfigure}{-1}\vspace{-2pt}\\%
  \foreach \env in {N17E073,N47W124}{%
    \subfloat[SMSE in \env\label{fig:\env_random_smse}]{%
      \resizebox{0.5\linewidth}{!}{%
        \includegraphics[width=0.5\linewidth,trim={3 3 3 3},clip]{metrics/\env/random/SMSE.pdf} %
      }%
    }%
  }\vspace{-5pt}\\%
  \foreach \env in {N17E073,N47W124}{%
    \subfloat[MLSS in \env\label{fig:\env_random_msll}]{%
      \resizebox{0.5\linewidth}{!}{%
        \includegraphics[width=0.5\linewidth,trim={3 3 3 3},clip]{metrics/\env/random/MSLL.pdf} %
      }%
    }%
  }%
  \caption{\textbf{SMSE and MSLL vs the number of samples in random sampling}.}\label{fig:metrics_random}\vspace{-10pt}%
\end{figure}

%% file: figures/tex/metrics_active.tex
\begin{figure}[t]
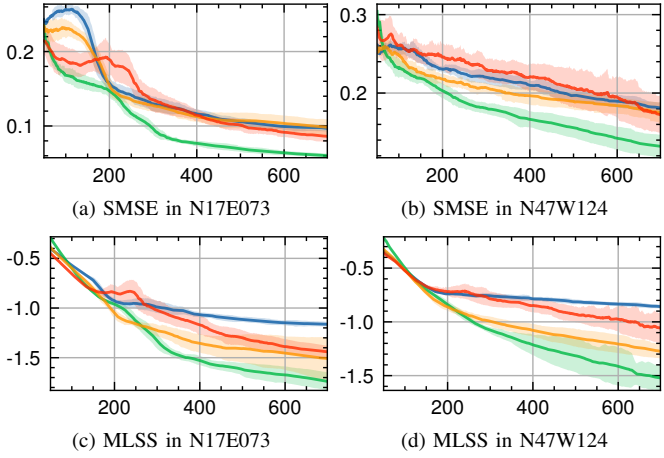

  \centering
  \foreach \env in {N17E073,N47W124}{%
    \subfloat[SMSE in \env\label{fig:\env_active_smse}]{%
      \resizebox{0.5\linewidth}{!}{%
        \includegraphics[width=0.5\linewidth,trim={3 3 3 3},clip]{metrics/\env/active/SMSE.pdf} %
      }%
    }%
  }\vspace{-5pt}\\%
  \foreach \env in {N17E073,N47W124}{%
    \subfloat[MLSS in \env\label{fig:\env_active_msll}]{%
      \resizebox{0.5\linewidth}{!}{%
        \includegraphics[width=0.5\linewidth,trim={3 3 3 3},clip]{metrics/\env/active/MSLL.pdf} %
      }%
    }%
  }%
  \caption{\textbf{SMSE and MSLL vs the number of samples in active sampling.}}\label{fig:metrics_active}\vspace{-10pt}%
\end{figure}

%% file: figures/tex/metrics_myopic.tex
\begin{figure}[t]
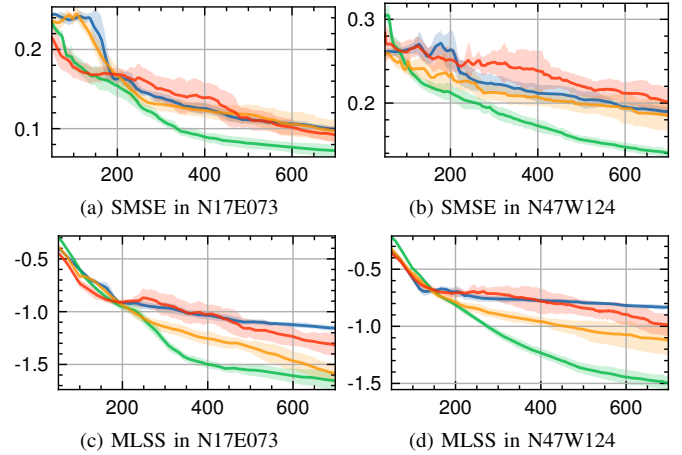

  \centering
  \foreach \env in {N17E073,N47W124}{%
    \subfloat[SMSE in \env\label{fig:\env_myopic_smse}]{%
      \resizebox{0.5\linewidth}{!}{%
        \includegraphics[width=0.5\linewidth,trim={3 3 3 3},clip]{metrics/\env/myopic/SMSE.pdf} %
      }%
    }%
  }\vspace{-5pt}\\%
  \foreach \env in {N17E073,N47W124}{%
    \subfloat[MLSS in \env\label{fig:\env_myopic_msll}]{%
      \resizebox{0.5\linewidth}{!}{%
        \includegraphics[width=0.5\linewidth,trim={3 3 3 3},clip]{metrics/\env/myopic/MSLL.pdf} %
      }%
    }%
  }%
  \caption{SMSE and MSLL of robotic information gathering experiments.}\label{fig:metrics_myopic}%
\end{figure}

%% file: figures/tex/predictions.tex
\begin{figure}[t]
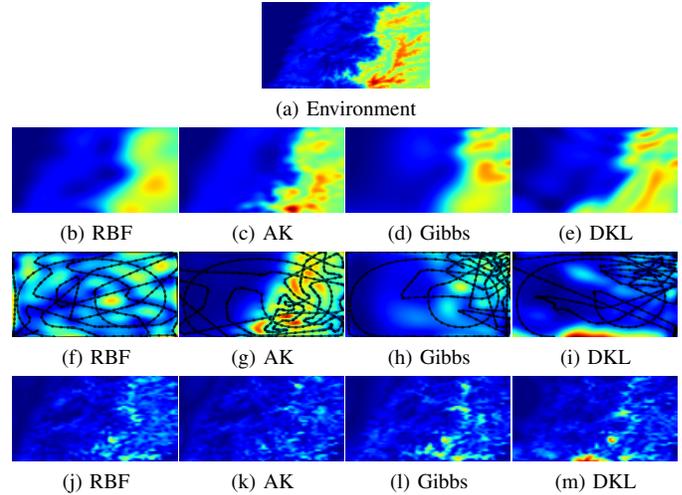

  \centering
  \foreach \env in {N17E073}{%
    \subfloat[Environment\label{fig:\env_prediction}]{%
      \resizebox{0.25\linewidth}{!}{%
    \includegraphics[width=0.25\linewidth]{predictions/\env.pdf}}}\vspace{-8pt}\\%
    \foreach \kernel in {RBF, AK, Gibbs, DKL}{%
      \subfloat[\kernel\label{fig:\kernel_prediction}]{%
        \resizebox{0.25\linewidth}{!}{%
    \includegraphics[width=\linewidth]{predictions/\env_\kernel_mean.pdf} }}}\vspace{-8pt}\\%
    \hspace{0.185\linewidth}
    \foreach \kernel in {RBF, AK, Gibbs, DKL}{%
      \subfloat[\kernel\label{fig:\kernel_uncertainty}]{%
        \resizebox{0.25\linewidth}{!}{%
    \includegraphics[width=\linewidth]{predictions/\env_\kernel_std.pdf} }}}\vspace{-8pt}\\%
    \hspace{0.185\linewidth}
    \foreach \kernel in {RBF, AK, Gibbs, DKL}{%
      \subfloat[\kernel\label{fig:\kernel_error}]{%
        \resizebox{0.25\linewidth}{!}{%
    \includegraphics[width=\linewidth]{predictions/\env_\kernel_error.pdf} }}}\vspace{-8pt}\\%
  }\vspace{8pt}
  \caption{\textbf{Snapshots of RIG with different kernels.} (b-e) show the prediction maps, (f-i) are the uncertainty maps with sampling paths, and (j-m) present the absolute error maps.}\label{fig:prediction}\vspace{-10pt}%
\end{figure}

%% file: figures/tex/field.tex
\begin{figure*}[t]
  \centering
  \subfloat[System\label{fig:field_system}]{%
    \resizebox{0.32\linewidth}{!}{%
  \includegraphics[width=\linewidth,height=0.6\linewidth]{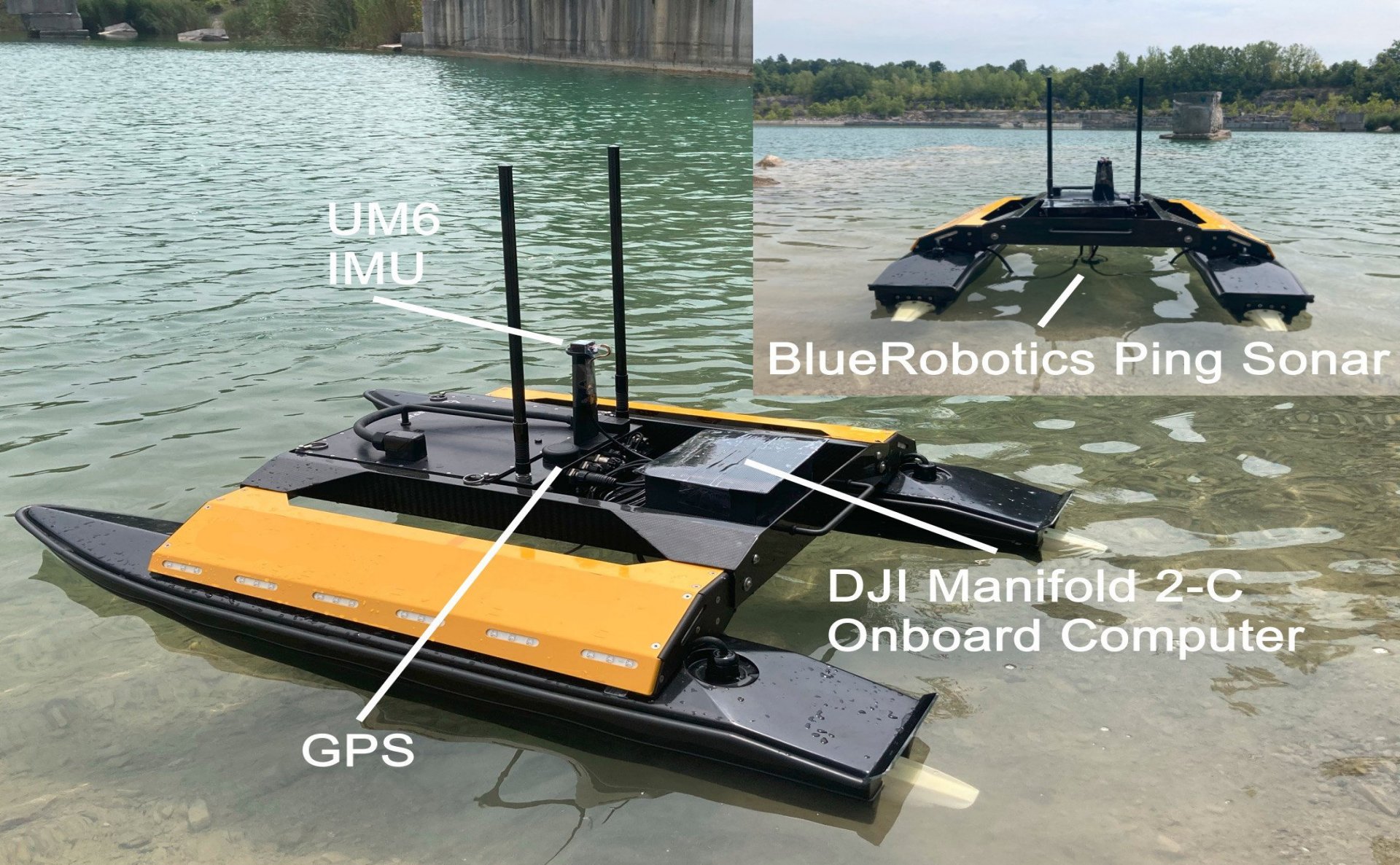} }}%
  \subfloat[Workspace\label{fig:field_workspace}]{%
    \resizebox{0.32\linewidth}{!}{%
      \begin{tikzpicture}%
        \node(a){\includegraphics[width=\linewidth,height=0.6\linewidth]{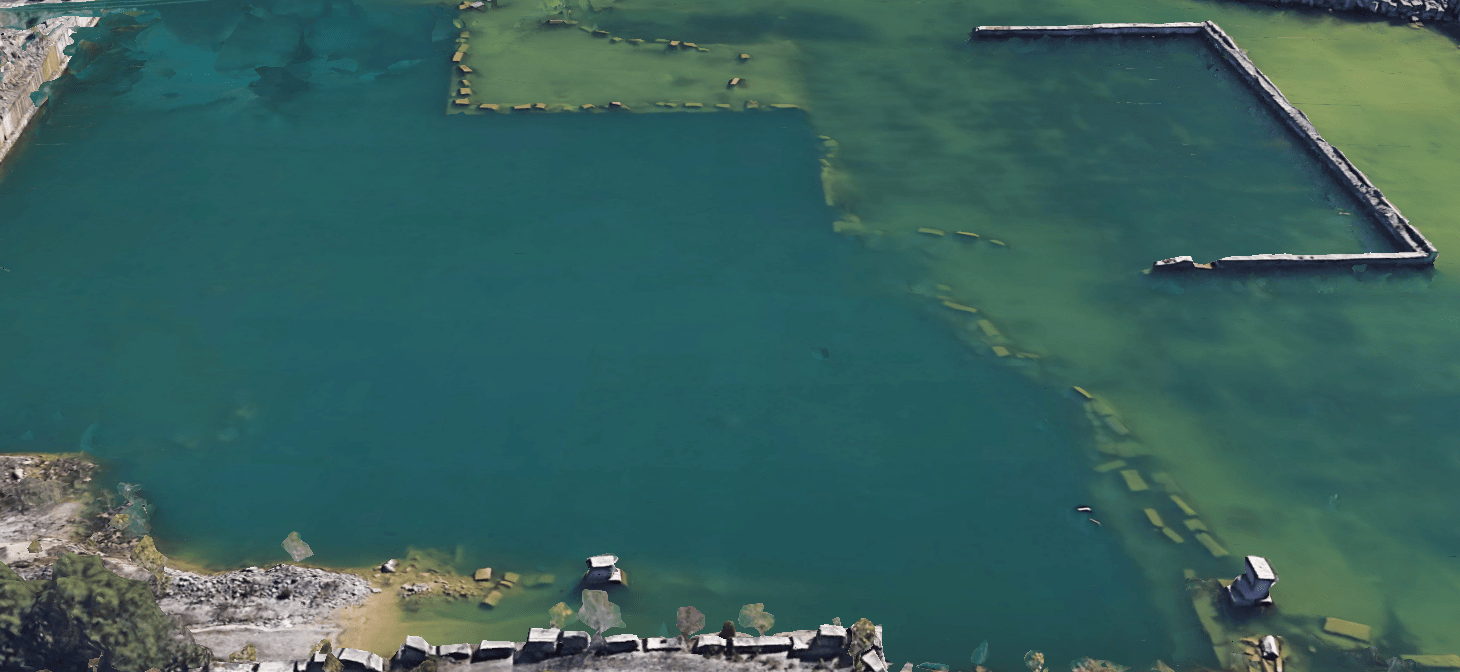} };%
        \node at(a.center)[draw,orange,line width=2,ellipse,minimum width=50,minimum height=50,rotate=0,yshift=-40pt,xshift=-180pt]{};%
        \node at(a.center)[draw,orange,line width=2,rectangle,minimum width=30,minimum height=50,rotate=0,yshift=50pt,xshift=35pt]{};%
        \draw [red,-stealth,line width=2](2,-3) -- (2,3);%
        \draw [green,-stealth,line width=2](2,-3) -- (-7,-3);%
        \node [white] at (2.5,0) {\Huge x};%
        \node [white] at (-2.5,-3.5) {\Huge y};%
  \end{tikzpicture}}}%
  \subfloat[Final Snapshot\label{fig:field_snapshot}]{%
    \resizebox{0.32\linewidth}{!}{%
      \begin{tikzpicture}
        \node(a){\includegraphics[width=\linewidth,height=0.6\linewidth]{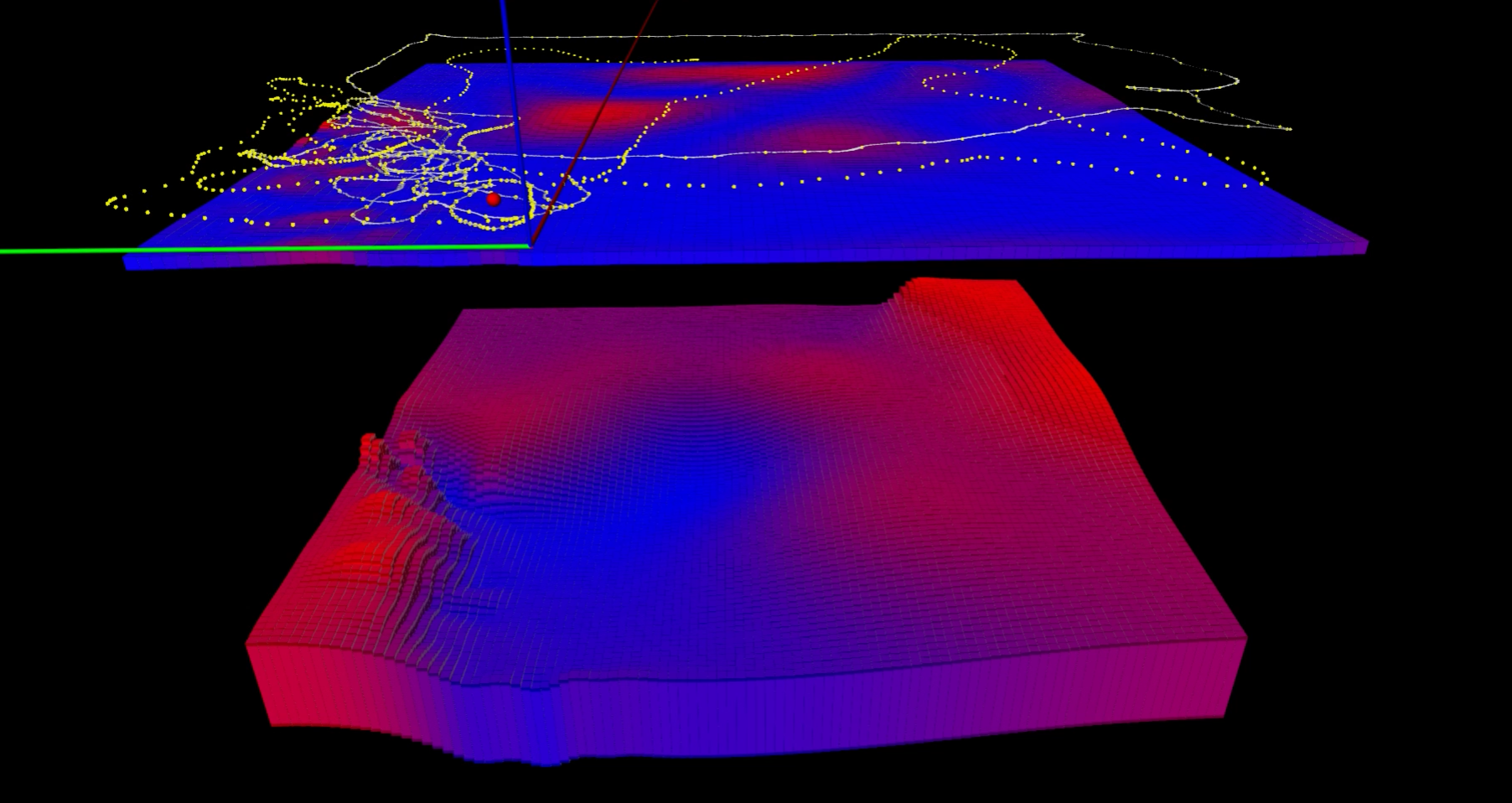}};%
        \node at(a.center)[draw,orange,line width=2,ellipse,minimum width=100,minimum height=100,rotate=0,yshift=-45pt,xshift=-110pt]{};%
        \node at(a.center)[draw,orange,line width=2,rectangle,minimum width=80,minimum height=80,rotate=0,yshift=10pt,xshift=80pt]{};%
        \draw [white,-stealth,line width=2](7,3.5) -- (6.5,3.7);%
        \node [white] at (7,3.2) {\LARGE Collected Data};%
        \draw [white,-stealth,line width=2](-2.5,2) -- (-3.1,2.7);%
        \node [white] at (-2.5,1.7) {\LARGE Informative Waypoint};%
        \draw [white,-stealth,line width=2](-2,4.3) -- (-2,3.9);%
        \node [white] at (-2,4.5) {\LARGE High Uncertainty};%
        \draw [white,-stealth,line width=2](4,2.5) -- (3.3,2.8);%
        \node [white] at (4.3,2.1) {\LARGE Low Uncertainty};%
        \draw [white,-stealth,line width=2](-5.2,0.2) -- (-4.8,-0.4);%
        \node [white] at (-5.2,0.5) {\LARGE High Elevation};%
        \draw [white,-stealth,line width=2](-0.5,-2) -- (-1,-1.5);%
        \node [white] at (-0.2,-2.5) {\LARGE Low Elevation};%
  \end{tikzpicture}}}%
  \caption{\textbf{An active elevation mapping field experiment}.}\label{fig:field_exp}\vspace{-10pt}
\end{figure*}
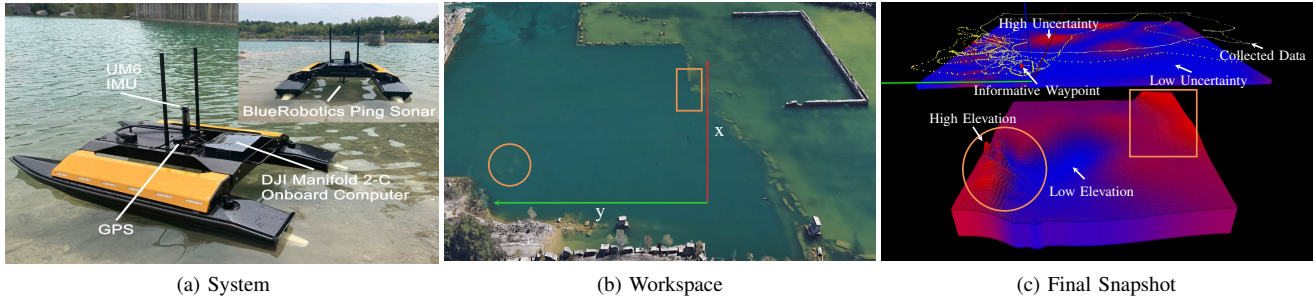

%% file: sections/6_conclusion.tex
\section{Conclusion and Future Work}\label{sec:6}
In this paper, we investigate the uncertainty quantification of probabilistic models, which is decisive for the performance of RIG but has received little attention. We present a family of nonstationary kernels called the Attentive Kernel, which is simple, robust, and can extend any stationary kernel to a nonstationary one. An extensive evaluation of elevation mapping tasks shows that AK provides better accuracy and uncertainty quantification than baselines. The improved uncertainty quantification guides the informative planning algorithms to collect more valuable samples around the high-error area, thus further reducing the prediction error. A field experiment demonstrates that AK enables an ASV to collect more samples in important sampling locations and capture the salient environmental features. 
The results indicate that misspecified probabilistic models affect the RIG performance profoundly. Future work 
includes further investigating 
the influence of \emph{outliers} and \emph{heteroscedastic noise} on RIG. Besides, a more principled training scheme of nonstationary kernels can be an essential future research direction.

\section{Acknowledgement} 
We acknowledge the support of NSF with grant numbers 1906694, 2006886, and 2047169. We are also grateful for the computational resources provided by the Amazon AWS Machine Learning Research Award. The constructive comments by the anonymous reviewers are greatly appreciated. We thank Durgakant Pushp and Mahmoud Ali for their help when conducting the field experiment. 

%% file: sections/7_appendix.tex
\subsection{Environments}\label{sec:appendix_envs}
\input{figures/tex/all_envs.tex}
\cref{fig:all_envs} shows all the environments used in the experiments. N17E073 consists of a flat part, a mountainous area, and a rocky region with many ridges. N43W080 presents a sharp elevation change at the north part while the lakebed is virtually flat. In N45W123, the environment has a complex upper part and a smoother lower part. There is also a river passing by from the middle. The right part of N47W124 varies drastically, and its left part is relatively flat.

\subsection{Benchmarking Tables}\label{sec:appendix_tables}
\input{tables/tex/all_random.tex}
\input{tables/tex/all_active.tex}
\input{tables/tex/all_myopic.tex}
To have a more evident quantitative comparison, we present all the benchmarking results in \cref{tab:random,tab:active,tab:rig}. Each number summarizes the metric curves by averaging the curves over the x-axis (\textit{i.e.}, the number of samples). This number indicates the averaged \emph{area under the curve}. A smaller area implies a faster drop in the curve. We can clearly see that AK has significantly better prediction accuracy (\textit{i.e.}, smaller SMSE, MSE, and MAE) and uncertainty quantification (\textit{i.e.}, smaller MSLL and NLPD).

\subsection{Sensitivity Analysis}\label{sec:appendix_sensitivity}
\input{figures/tex/sensitivity_m.tex}
\input{figures/tex/sensitivity_h.tex}
\input{figures/tex/sensitivity_min.tex}
\input{figures/tex/sensitivity_max.tex}
\cref{fig:sensitivity_m} presents the results of sensitivity analysis of the number of base kernels $M$, which should be larger than $2$. Increasing $M$ brings better performance, albeit with a diminishing return and higher computational complexity. Choosing a number in the range of $[5, 10]$ is a good tradeoff between performance and computational efficiency. \cref{fig:sensitivity_h} shows that the AK is not sensitive to the number of hidden units in the neural network as long as $H$ is not too small. In \cref{fig:sensitivity_min}, smaller minimum lengthscales yield better performance with a diminishing return. The blue line and the green line are overlapped, which means that the advantage is negligible when choosing a minimum lengthscale smaller than $0.01$. Therefore, $0.01$ is an appropriate choice. As shown in \cref{fig:sensitivity_max}, the AK is robust to the choice of the maximum lengthscale as long as it is not too small. If the inputs are normalized to $[-1, 1]$, choosing a value in the range $[0.5, 1.0]$ is reasonable.

\subsection{Ablation Study}\label{sec:appendix_ablation}
\input{figures/tex/ablation.tex}
We compare four variants of the attentive kernel in the random sampling experiments for the ablation study. \textit{Full} means the AK presented in the paper, \textit{Weight} represents the AK with only lengthscale selection, \textit{Mask} stands for instance selection alone, and \textit{NNx2} uses two separated neural networks. The results show that lengthscale selection is necessary, and dropping it decreases the performance significantly (see the \textit{Mask} line). We do not observe a significant performance advantage from instance selection. Using two separate neural networks for the weighting function and the membership function does not provide an improvement but deteriorates the uncertainty quantification in one environment (\textit{i.e.}, N43W080).

\subsection{Overfitting Analysis}\label{sec:appendix_overfitting}
\input{figures/tex/overfitting.tex}
\cref{fig:overfitting} shows the training and test MSLL. We have repeated the analysis in other environments, but only two representative environments are presented here for compactness. In some environments, as shown in \cref{fig:volcano_train_msll,fig:volcano_test_msll}, the AK is fairly robust while the Gibbs kernel and DKL show a clear overfitting trend. However, all the nonstationary kernels suffer from overfitting in some environments such as N17E073. The difference seems to be related to the sharp changes in the environment.


%% file: figures/tex/all_envs.tex
\begin{figure}[htb]%
  \centering
  \foreach \env in {N17E073,N43W080,N45W123,N47W124}{%
    \subfloat[\env]{%
    \includegraphics[width=0.25\linewidth]{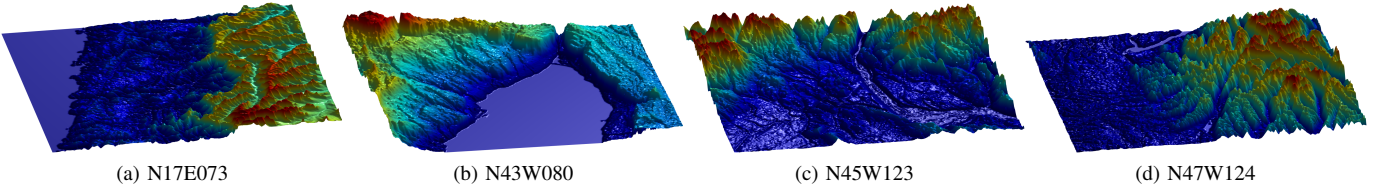}}}%
    \caption{\textbf{The four environments used in the elevation mapping tasks}. Red means high elevation, and blue represents low elevation. }\label{fig:all_envs}\vspace{-10pt}%
\end{figure}

%% file: tables/tex/all_random.tex
\begin{table}[htb]
  \caption{Random sampling performance.}\label{tab:random}
  \centering
  \scriptsize
  \begin{tabular}{
      ll
      S[table-format=1.2(3)e-1]
      S[table-format=-1.2(3)e-1]
      S[table-format=1.2(3)]
      S[table-format=1.2(3)e1]
      S[table-format=1.2(3)e1]
    }
    \toprule
    {Environment} & {Method} & {SMSE$\downarrow^{1}_{0}$} & {MSLL$\downarrow^{0}$} & {NLPD$\downarrow$} & {RMSE$\downarrow_{0}$} & {MAE$\downarrow_{0}$} \\
    \midrule
    N17E073   & RBF   &            1.33(3)e-01 &             -9.9(1)e-01 &            4.59(1)e+00 &            2.33(3)e+01 &            1.69(3)e+01 \\
              & AK    &  \B 1.11(4)e-01 &  \B -1.24(1)e+00 &  \B 4.34(1)e+00 &  \B 2.13(4)e+01 &  \B 1.50(2)e+01 \\
              & Gibbs &            1.33(1)e-01 &            -1.09(2)e+00 &            4.50(3)e+00 &            2.33(9)e+01 &            1.66(4)e+01 \\
              & DKL   &            1.37(6)e-01 &             -9.7(3)e-01 &            4.62(3)e+00 &            2.37(5)e+01 &            1.68(4)e+01 \\
              \midrule
    N43W080   & RBF   &            7.1(3)e-02 &            -1.43(2)e+00 &            3.87(2)e+00 &            1.23(3)e+01 &           8.13(6)e+00 \\
              & AK    &  \B 6.0(5)e-02 &  \B -1.69(6)e+00 &  \B 3.62(6)e+00 &  \B 1.11(5)e+01 &  \B 7.0(2)e+00 \\
              & Gibbs &            7.2(4)e-02 &            -1.48(6)e+00 &            3.83(6)e+00 &            1.25(5)e+01 &            8.3(3)e+00 \\
              & DKL   &            6.6(8)e-02 &            -1.49(4)e+00 &            3.81(4)e+00 &            1.19(7)e+01 &            7.5(3)e+00 \\
              \midrule
    N45W123   & RBF   &            1.65(7)e-01 &             -9.4(3)e-01 &            4.37(3)e+00 &            1.97(4)e+01 &            1.28(3)e+01 \\
              & AK    &  \B 1.41(6)e-01 &  \B -1.28(2)e+00 &  \B 4.03(2)e+00 &  \B 1.80(4)e+01 &  \B 1.15(2)e+01 \\
              & Gibbs &             1.8(1)e-01 &            -1.08(1)e+00 &            4.24(2)e+00 &            2.07(7)e+01 &            1.34(2)e+01 \\
              & DKL   &             2.0(1)e-01 &             -9.1(1)e-01 &            4.41(1)e+00 &            2.18(7)e+01 &            1.42(6)e+01 \\
              \midrule
    N47W123   & RBF   &            2.26(7)e-01 &             -7.2(1)e-01 &            4.77(1)e+00 &            2.77(4)e+01 &            1.97(2)e+01 \\
              & AK    &  \B 1.90(5)e-01 &  \B -1.06(1)e+00 &  \B 4.43(1)e+00 &  \B 2.53(3)e+01 &  \B 1.77(2)e+01 \\
              & Gibbs &            2.21(8)e-01 &             -7.7(4)e-01 &            4.72(5)e+00 &            2.74(5)e+01 &            1.94(3)e+01 \\
              & DKL   &            2.34(8)e-01 &             -7.1(2)e-01 &            4.78(2)e+00 &            2.82(5)e+01 &            1.98(3)e+01 \\
              \bottomrule
  \end{tabular}\vspace{-10pt}%
\end{table}

%% file: tables/tex/all_active.tex
\begin{table}[ht]
  \caption{Active sampling performance.}\label{tab:active}
  \centering
  \scriptsize
  \begin{tabular}{
      ll
      S[table-format=1.2(3)e-1]
      S[table-format=-1.2(3)e-1]
      S[table-format=1.2(3)]
      S[table-format=1.2(3)e1]
      S[table-format=1.2(3)e1]
    }
    \toprule
    {Environment} & {Method} & {SMSE$\downarrow^{1}_{0}$} & {MSLL$\downarrow^{0}$} & {NLPD$\downarrow$} & {RMSE$\downarrow_{0}$} & {MAE$\downarrow_{0}$} \\
    \midrule
    N17E073   & RBF   &            1.41(4)e-01 &             -9.8(2)e-01 &            4.61(2)e+00 &            2.38(3)e+01 &            1.70(3)e+01 \\
              & AK    &  \B 1.01(2)e-01 &  \B -1.32(4)e+00 &  \B 4.36(2)e+00 &  \B 2.00(2)e+01 &  \B 1.43(2)e+01 \\
              & Gibbs &            1.37(6)e-01 &            -1.20(8)e+00 &            4.59(3)e+00 &            2.35(6)e+01 &            1.72(5)e+01 \\
              & DKL   &            1.33(7)e-01 &            -1.09(5)e+00 &            4.59(3)e+00 &            2.32(6)e+01 &            1.62(5)e+01 \\
              \midrule
    N43W080   & RBF   &            7.8(2)e-02 &            -1.41(1)e+00 &            3.96(1)e+00 &            1.28(1)e+01 &            9.0(1)e+00 \\
              & AK    &  \B 5.1(2)e-02 &  \B -1.72(2)e+00 &  \B 3.74(3)e+00 &  \B 1.02(2)e+01 &  \B 6.9(1)e+00 \\
              & Gibbs &            8.0(6)e-02 &            -1.48(5)e+00 &            3.98(6)e+00 &            1.31(6)e+01 &            9.8(4)e+00 \\
              & DKL   &              7(1)e-02 &             -1.6(1)e+00 &             3.9(1)e+00 &             1.2(1)e+01 &            8.2(6)e+00 \\
              \midrule
    N45W123   & RBF   &            1.47(4)e-01 &             -9.7(1)e-01 &            4.36(1)e+00 &            1.85(2)e+01 &            1.23(2)e+01 \\
              & AK    &  \B 1.08(3)e-01 &  \B -1.55(4)e+00 &  \B 4.16(2)e+00 &  \B 1.57(3)e+01 &  \B 1.14(3)e+01 \\
              & Gibbs &            1.29(6)e-01 &            -1.48(5)e+00 &            4.30(2)e+00 &            1.73(4)e+01 &            1.28(2)e+01 \\
              & DKL   &             1.6(1)e-01 &            -1.18(4)e+00 &            4.35(3)e+00 &            1.91(7)e+01 &            1.35(4)e+01 \\
              \midrule
    N47W124   & RBF   &            2.15(5)e-01 &             -7.5(1)e-01 &            4.75(1)e+00 &            2.70(3)e+01 &            1.90(3)e+01 \\
              & AK    &  \B 1.78(8)e-01 &  \B -1.09(7)e+00 &  \B 4.56(1)e+00 &  \B 2.45(6)e+01 &  \B 1.75(3)e+01 \\
              & Gibbs &            2.04(6)e-01 &             -9.9(5)e-01 &            4.71(2)e+00 &            2.63(4)e+01 &            1.86(3)e+01 \\
              & DKL   &             2.2(1)e-01 &             -8.1(5)e-01 &            4.76(5)e+00 &            2.75(9)e+01 &            1.94(5)e+01 \\
              \bottomrule
  \end{tabular}\vspace{-10pt}%
\end{table}

%% file: tables/tex/all_myopic.tex
\begin{table}[htb]
  \caption{Robotic information gathering performance.}\label{tab:rig}
  \centering
  \scriptsize
  \begin{tabular}{
      ll
      S[table-format=1.2(3)e-1]
      S[table-format=-1.2(3)e-1]
      S[table-format=1.2(3)]
      S[table-format=1.2(3)e1]
      S[table-format=1.2(3)e1]
    }
    \toprule
    {Environment} & {Method} & {SMSE$\downarrow^{1}_{0}$} & {MSLL$\downarrow^{0}$} & {NLPD$\downarrow$} & {RMSE$\downarrow_{0}$} & {MAE$\downarrow_{0}$} \\
    \midrule
    N17E073   & RBF   &            1.45(3)e-01 &             -9.7(2)e-01 &            4.63(2)e+00 &            2.42(2)e+01 &            1.73(2)e+01 \\
              & AK    &  \B 1.14(4)e-01 &  \B -1.27(3)e+00 &  \B 4.41(4)e+00 &  \B 2.14(4)e+01 &  \B 1.51(2)e+01 \\
              & Gibbs &            1.43(7)e-01 &            -1.16(4)e+00 &            4.61(4)e+00 &            2.40(7)e+01 &            1.76(6)e+01 \\
              & DKL   &            1.38(9)e-01 &            -1.01(6)e+00 &            4.61(4)e+00 &            2.38(8)e+01 &            1.67(6)e+01 \\
              \midrule
    N43W080   & RBF   &            7.7(4)e-02 &            -1.40(2)e+00 &            3.94(2)e+00 &            1.27(3)e+01 &             8.8(2)e+00 \\
              & AK    &  \B 6.6(2)e-02 &  \B -1.64(4)e+00 &  \B 3.78(3)e+00 &  \B 1.14(2)e+01 &  \B 7.69(9)e+00 \\
              & Gibbs &            7.6(9)e-02 &            -1.50(5)e+00 &            3.91(7)e+00 &            1.25(7)e+01 &             9.0(6)e+00 \\
              & DKL   &            7.0(1)e-02 &            -1.56(7)e+00 &            3.85(6)e+00 &            1.19(8)e+01 &             8.1(6)e+00 \\
              \midrule
    N45W123   & RBF   &            1.60(6)e-01 &             -9.3(2)e-01 &            4.39(2)e+00 &            1.93(4)e+01 &            1.29(2)e+01 \\
              & AK    &  \B 1.32(6)e-01 &  \B -1.43(4)e+00 &  \B 4.15(3)e+00 &  \B 1.71(4)e+01 &  \B 1.21(3)e+01 \\
              & Gibbs &            1.38(7)e-01 &            -1.34(4)e+00 &            4.30(3)e+00 &            1.79(5)e+01 &            1.32(4)e+01 \\
              & DKL   &             1.7(2)e-01 &            -1.06(8)e+00 &            4.41(6)e+00 &            1.99(9)e+01 &            1.40(6)e+01 \\
              \midrule
    N47W124   & RBF   &            2.23(6)e-01 &             -7.4(1)e-01 &            4.76(1)e+00 &            2.75(3)e+01 &            1.94(2)e+01 \\
              & AK    &  \B 1.85(4)e-01 &  \B -1.10(3)e+00 &  \B 4.48(3)e+00 &  \B 2.50(3)e+01 &  \B 1.79(3)e+01 \\
              & Gibbs &            2.12(8)e-01 &             -9.0(5)e-01 &            4.73(3)e+00 &            2.69(5)e+01 &            1.91(2)e+01 \\
              & DKL   &            2.36(6)e-01 &             -7.7(4)e-01 &            4.78(3)e+00 &            2.83(3)e+01 &            1.99(4)e+01 \\
              \bottomrule
  \end{tabular}\vspace{-10pt}%
\end{table}

%% file: figures/tex/sensitivity_m.tex
\begin{figure}[htb]
  \centering
  \subfloat{\includegraphics[width=0.6\linewidth]{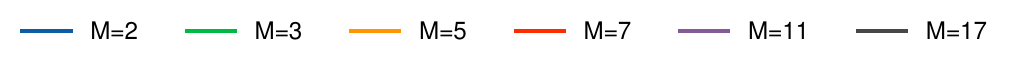}}%
  \addtocounter{subfigure}{-1}\vspace{-10pt}\\%
  \foreach \env in {N17E073,N43W080,N45W123,N47W124}{%
    \subfloat[MSLL in \env]{%
    \includegraphics[width=0.25\linewidth]{./sensitivity/m/\env/MSLL.pdf}}}
    \caption{\textbf{Sensitivity analysis of the choice of $M$}.}\label{fig:sensitivity_m}\vspace{-10pt}%
\end{figure}

%% file: figures/tex/sensitivity_h.tex
\begin{figure}[htb]
  \centering
  \subfloat{\includegraphics[width=0.6\linewidth]{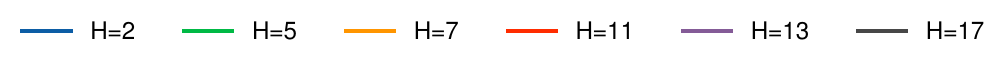}}%
  \addtocounter{subfigure}{-1}\vspace{-10pt}\\%
  \foreach \env in {N17E073,N43W080,N45W123,N47W124}{%
    \subfloat[MSLL in \env]{%
    \includegraphics[width=0.25\linewidth]{./sensitivity/h/\env/MSLL.pdf}}}
    \caption{\textbf{Sensitivity analysis of the choice of $H$}.}\label{fig:sensitivity_h}\vspace{-10pt}%
\end{figure}

%% file: figures/tex/sensitivity_min.tex
\begin{figure}[htb]
  \centering
  \subfloat{\includegraphics[width=0.9\linewidth]{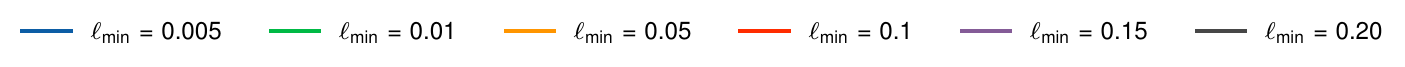}}%
  \addtocounter{subfigure}{-1}\vspace{-10pt}\\%
  \foreach \env in {N17E073,N43W080,N45W123,N47W124}{%
    \subfloat[MSLL in \env]{%
    \includegraphics[width=0.25\linewidth]{./sensitivity/min/\env/MSLL.pdf}}}
    \caption{\textbf{Sensitivity analysis of the choice of $\ell_{\text{min}}$}.}\label{fig:sensitivity_min}%
\end{figure}

%% file: figures/tex/sensitivity_max.tex
\begin{figure}[htb]
  \centering
  \subfloat{\includegraphics[width=0.9\linewidth]{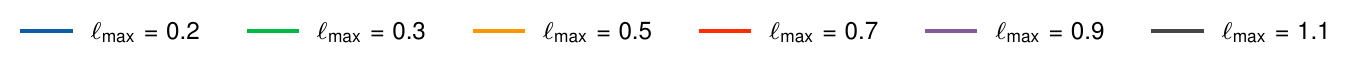}}%
  \addtocounter{subfigure}{-1}\vspace{-10pt}\\%
  \foreach \env in {N17E073,N43W080,N45W123,N47W124}{%
    \subfloat[MSLL in \env]{%
    \includegraphics[width=0.25\linewidth]{./sensitivity/max/\env/MSLL.pdf}}}
    \caption{\textbf{Sensitivity analysis of the choice of $\ell_{\text{max}}$}.}\label{fig:sensitivity_max}%
\end{figure}

%% file: figures/tex/ablation.tex
\begin{figure}[htb]%
  \centering
  \subfloat{\includegraphics[width=0.45\linewidth]{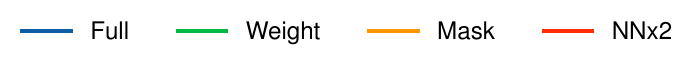}}%
  \addtocounter{subfigure}{-1}\vspace{-10pt}\\%
  \foreach \env in {N17E073,N43W080,N45W123,N47W124}{%
    \subfloat[MSLL in \env]{%
    \includegraphics[width=0.25\linewidth]{./ablation/\env/MSLL.pdf}}}%
    \caption{\textbf{Results of the four variants in the ablation study}.}\label{fig:ablation}\vspace{-10pt}%
\end{figure}

%% file: figures/tex/overfitting.tex
\begin{figure}[htb]%
  \centering
  \subfloat{\includegraphics[width=0.45\linewidth]{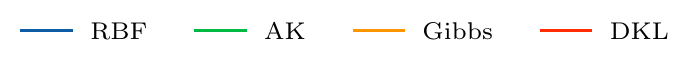}}%
  \addtocounter{subfigure}{-1}\vspace{-10pt}\\%
  \subfloat[Training MSLL in Volcano\label{fig:volcano_train_msll}]{%
    \resizebox{0.25\linewidth}{!}{%
  \includegraphics[width=\linewidth,height=0.6\linewidth]{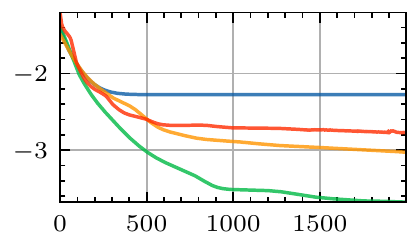} }}%
  \subfloat[Test MSLL in Volcano\label{fig:volcano_test_msll}]{%
    \resizebox{0.25\linewidth}{!}{%
  \includegraphics[width=\linewidth,height=0.6\linewidth]{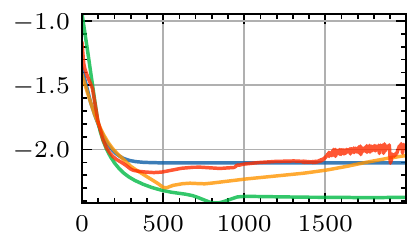} }}%
  \subfloat[Training MSLL in N17E073]{%
    \resizebox{0.25\linewidth}{!}{%
  \includegraphics[width=\linewidth,height=0.6\linewidth]{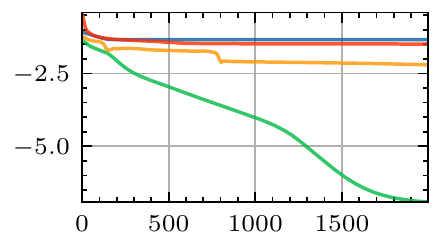} }}%
  \subfloat[Test MSLL in N17E073]{%
    \resizebox{0.25\linewidth}{!}{%
  \includegraphics[width=\linewidth,height=0.6\linewidth]{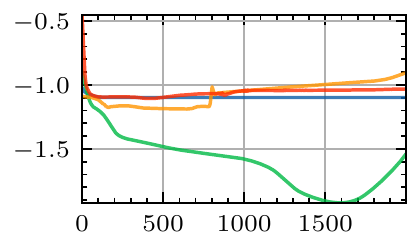}}}\\%
  \caption{\textbf{Results of the overfitting analysis} in the \textit{Volcano} environment introduced in \cref{fig:volcano_env} and N17E073.}\label{fig:overfitting}%
\end{figure}